\documentclass[11pt,a4paper,twoside]{article}

\usepackage{graphicx,bm,url,color,bbm,latexsym}
\usepackage{amsfonts,amssymb,amsthm,amsmath,enumitem}
\usepackage[english]{babel}
\usepackage{natbib}
\usepackage{float} 
\usepackage[bookmarks=false]{hyperref} 
\usepackage{mathrsfs} 
\usepackage{multirow}

\newcommand{\bK}{\mathbf K}
\newcommand{\bx}{\mathbf x}
\newcommand{\bX}{\mathbf X}
\newcommand{\by}{\mathbf y}
\newcommand{\bY}{\mathbf Y}

\newcommand{\bSigma}{\boldsymbol\Sigma}

\newcommand{\bbN}{\mathbb N}

\newcommand{\bbR}{\mathbb R}

\newcommand{\calM}{\mathcal M}
\newcommand{\calP}{\mathcal P}

\newcommand{\calX}{\mathcal X}

\newcommand{\scrC}{\mathscr C}
\newcommand{\scrH}{\mathscr H}

\def \MMD{\text{\rm MMD}}
\def \WCSS{\text{\rm WCSS}}

\newcommand{\mbf}[1]{\mathbf{#1}}

\usepackage{mathtools}
\DeclarePairedDelimiter\floor{\lfloor}{\rfloor}

\title{\textbf{\Large Kernel \(K\)-means clustering \\ of distributional data}}
\author{
Amparo Ba\'illo$^1$\footnote{Corresponding author: A. Ba\'illo (amparo.baillo@uam.es)}\,, Jos\'{e} R. Berrendero$^1$\, and Mart\'{\i}n S\'{a}nchez-Signorini$^{2}$ \\
{\small{$^{1}$ Departamento de Matem\'{a}ticas, Universidad Aut\'{o}noma de Madrid, 28049 Madrid (SPAIN)}}\\
{\small{$^{2}$ Escuela de Doctorado, Universidad Aut\'{o}noma de Madrid, 28049 Madrid (SPAIN)}}
}
\date{}

\providecommand{\keywords}[1]{\textbf{Keywords:} #1}

\begin{document}

\maketitle

\begin{abstract}
\noindent
We consider the problem of clustering a sample of probability distributions from a random distribution on \(\bbR^p\).
Our proposed partitioning method makes use of a symmetric, positive-definite kernel \(k\) and its associated reproducing kernel Hilbert space (RKHS) \(\scrH\).
By mapping each distribution to its corresponding kernel mean embedding in \(\scrH\), we obtain a sample in this RKHS where we carry out the \(K\)-means clustering procedure, which provides an unsupervised classification of the original sample.
The procedure is simple and computationally feasible even for dimension \(p>1\).
The simulation studies provide insight into the choice of the kernel and its tuning parameter.
The performance of the proposed clustering procedure is illustrated on a collection of Synthetic Aperture Radar (SAR) images.
\end{abstract}




\section{Introduction} \label{Section.Introduction}

This work focuses on the unsupervised classification of {\em Distributional Data} (see \cite{BritoDias22}), which can be considered a subarea of Functional Data Analysis (see \cite{AriasCastroQiao25}, \cite{DensitiesAsDataObjects}).
Distributional data are realizations of a random distribution, here denoted by \(F\) or \(P_F\) indistinctly, that is, a random object taking values in a space of probability distributions.
A distributional observation can arise, for instance, as a summary of a very large sample (e.g., as produced by a wearable device), or of a more complex random object (e.g., an image); see \cite{MoraLopezMora15}, \cite{DensitiesAsDataObjects}.
Samples of probability distributions are also obtained from histogram-valued random variables, giving rise to a type of symbolic data (see, e.g., \cite{Billard06}).

A popular tool for analyzing samples of (usually univariate) probability distributions is to use the geometry provided by the Wasserstein metric (see, e.g., \cite{Bigot20}, \cite{Irpino_et_al_14}, \cite{OkanoImaizumi24}). Also employed in the context of distributional data is the \(L^p\) distance, for \(1\leq p\leq\infty\), between the cumulative distribution functions (cdfs) or the densities (see \cite{MoraLopezMora15}) or the Kullback-Leibler divergence (see \cite{PetersenMuller16} and references therein). However, these discrepancies are generally not feasible nor practical when dealing with multivariate distributions. To overcome this difficulty, \cite{Gachon_et_al_25} have proposed a quantization of the distributions as a means to reduce their complexity.

In this work we consider the problem of clustering a sample of iid probability distributions \(F_i\), \(i=1,\ldots,n\), from the random (possibly multivariate) distribution \(F\).
We assume that the support of any realization of \(F\) is contained in a common set \(\calX \subseteq\bbR^p\).
From now on we denote by \(\calM(\calX)\) the collection of probability distributions whose support is contained in \(\calX\).
There have been different proposals to classify (mainly univariate) distributions with the \(K\)-means procedure. The differences are generally due to the metric or dissimilarity chosen to quantify the discrepancy between the distributional data; or to adaptive variations of the original \(K\)-means algorithm.
For continuous distributions \cite{EP_MEANS15} used the 1-Wasserstein distance between the cdfs, \cite{MoraLopezMora15} and \cite{Zhu21BankCard} employed the two-sample Kolmogorov-Smirnov test statistic, \cite{Irpino_et_al_17} proposed a fuzzy algorithm based on the 2-Wasserstein metric.
Regarding the \(K\)-means procedure applied to multivariate distributions, \cite{Irpino_et_al_14} employ the 2-Wasserstein distance of the unidimensional marginals to measure the dissimilarity between histograms; \cite{VerdinelliWasserman19} propose to use a hybrid Wasserstein distance; \cite{Barrio_et_al_19} develop a trimmed \(K\)-means procedure using the 2-Wasserstein metric and assume that for dimension \(p\geq 2\) the sampled probabilities are in the same location-scatter family.
Other types of clustering methodologies proposed for distributional data (i) are hierarchical (\cite{Applegate_et_al_11}, \cite{IrpinoVerde06}); (ii) rely on using a basis expansion to reduce the dimension of the distributions (\cite{MontanariCalo13}); (iii) perform \(K\)-prototypes algorithms where the prototypes are not the means (\cite{ChenHung14}); (iv) carry out iterative reassignments of the data to the clusters with a bound on the cluster width (\cite{VovanPhamgia10}); (v) are model-based (\cite{Vrac_et_al_12}).

Here, given a kernel \(k\), we propose to (implicitly) map the observed distributions \(F_i\), via their kernel mean embedding \(\mu_{F_i}\), into the reproducing kernel Hilbert space (RKHS) \(\scrH\) associated to \(k\). Then, the \(K\)-means algorithm is carried out on the transformed sample \(\mu_{F_1},\ldots,\mu_{F_n}\) in \(\scrH\), by using the maximum mean discrepancy as the distance between the original infinite-dimensional data \(F_1,\ldots,F_n\).
The paper is structured as follows.
In Section~\ref{Section.KKmeansDD} we describe in detail the background and the clustering algorithm, as well as several technical details regarding its practical implementation.
Section~\ref{Section.Simulations} summarizes the results of two simulation studies, with uni- and bi-dimensional probability distributions, respectively.
Section~\ref{Section.RealData} illustrates the performance of kernel \(K\)-means clustering on real probability distributions derived from SAR images, both in the uni- and bi-dimensional setting. Finally, Section~\ref{Section.Conclusions} describes the main contributions of this work and provides conclusions.


\section{Kernel \(K\)-means clustering of distributional data} \label{Section.KKmeansDD}

\subsection{The framework}

Mapping or embedding the sample of distributions into a Hilbert space may enable the application of multivariate statistical techniques to this type of infinite-dimensional data (see, e.g., \cite{PetersenMuller16}, \cite{AriasCastroQiao25}).
We propose a kernel \(K\)-means classification procedure based on directly embedding the distributions into a RKHS. Note that we do not embed the \(p\)-dimensional observations from the realizations of \(F\) into the RKHS (as it is done in kernel \(K\)-means clustering of multidimensional observations; see \cite{Dhillon_et_al_04}), since it is an unnecessary approach in the context of distributional data.
Given a positive-definite, symmetric kernel \(k:\calX\times\calX\to\bbR\), we consider the kernel mean embedding (KME)
\begin{equation} \label{KmeanEmbed}
\mu_P(\cdot) := \int_\calX k(\bx,\cdot) \, dP(\bx)
\end{equation}
of each probability distribution \(P\in\calM(\calX)\) (see, e.g., \cite{BOOK_KernelMeanEmbedDistrib}).
If
\begin{equation} \label{KmeanEmbed_in_RKHS}
E_{\bX\sim P}(\sqrt{k(\bX,\bX)}) <\infty,
\end{equation}
then \(\mu_P\in\scrH\), where \(\scrH\) is the reproducing kernel Hilbert space (RKHS) corresponding to the kernel \(k\). From now on we assume that \eqref{KmeanEmbed_in_RKHS} holds for the chosen kernel and for the distributions considered in this work. In particular, we assume that, for almost every realization \(F(\omega)\) of the random distribution \(F\) and for \(\bX\sim F(\omega)\), \eqref{KmeanEmbed_in_RKHS} is fulfilled. If \(k\) is bounded, as in the case of the Gaussian and Laplace kernels, \eqref{KmeanEmbed_in_RKHS} automatically holds for any distribution \(P\).

\subsection{\(K\)-means clustering of kernel mean embeddings}

To carry out any clustering procedure on a sample, it is necessary to have a metric or a dissimilarity measure, \(d\), between the sampled objects. After mapping the probability distributions to their corresponding kernel mean embedding, it is natural to use the maximum mean discrepancy (MMD) as the distance between them. Given two probability distributions \(P\) and \(Q\) on \(\calX\), the MMD is the distance in \(\scrH\) between their embeddings, \(\MMD(\scrH,P,Q)=\|\mu_P-\mu_Q\|_{\scrH}\), which satisfies
\begin{equation} \label{Squared_MMD}
\|\mu_P-\mu_Q\|_{\scrH}^2 = E_{\bX,\bX'} k(\bX,\bX') + E_{\bY,\bY'} k(\bY,\bY') -2 E_{\bX,\bY} k(\bX,\bY),
\end{equation}
where \(\bX,\bX'\sim P\) and \(\bY,\bY'\sim Q\) are independent r.v.'s (see \cite{Gretton_et_al_12a}).

For the distributional sample \(F_i\), \(i=1,\ldots,n\), we define the kernel \(K\)-means as the elements \(G_j\in\calM(\calX)\), \(j=1,\ldots,K\), minimizing
\begin{equation} \label{Empirical_Objective_Fn}
\sum_{i=1}^n \left( \min_{j=1,\ldots,K} \MMD(\scrH,F_i,G_j)^2 \right).
\end{equation}
The centroids \(G_1,\ldots,G_K\) are used to partition the sample into \(K\) clusters, \(C_1,\ldots,C_K\) respectively. The \(j\)-th cluster \(C_j\) is formed by the \(n_j\) elements \(F_{i_{1}},\ldots,F_{i_{n_j}}\) of the sample which are closer (with respect to the MMD of \(\scrH\)) to the \(j\)-th centroid \(G_j\) than to any other centroid.

We denote by \(\mu_F,\mu_{F_1},\ldots,\mu_{F_n}\) the kernel mean embeddings \eqref{KmeanEmbed} of the probability measures given by \(F,F_1,\ldots,F_n\) respectively.
Our kernel \(K\)-means clustering proposal of the sample of distributions \(F_i\), \(i=1,\ldots,n\), consists in carrying out the \(K\)-means algorithm on the embedded sample \(\mu_{F_1},\ldots,\mu_{F_n}\) with respect to the metric \(\|\;\|_\scrH\).
To this end, observe that \(\bar F=\sum_{i=1}^n F_i/n\), the sample mean of the distribution functions, is also a probability distribution satisfying \eqref{KmeanEmbed_in_RKHS} and, consequently, can be embedded into \(\mu_{\bar F}\in\scrH\). Since \(\mu_{\bar F}\) is the sample mean of embeddings \(\mu_{F_1},\ldots,\mu_{F_n}\), then it
is actually the element \(\nu\in\scrH\) minimizing \(\sum_{i=1}^n \|\mu_{F_i}-\nu\|^2_{\scrH}\).
Therefore, the kernel \(K\)-means minimizing \eqref{Empirical_Objective_Fn} can be expressed as \(G_j=\bar F_j=\sum_{i=1}^{n_j} F_{ji}/{n_j}\), \(j=1,\ldots,K\), the sample mean of the distributions in cluster \(C_j\).

\subsection{The algorithm} \label{Subsection.Algorithm}

Let \(F_i\), \(i=1,\ldots,n\) be the sample of probability distributions to be classified into \(K\geq 1\) clusters. Let \(k\) be the kernel. The kernel \(K\)-means clustering algorithm groups the distributional data into \(K\) subgroups, \(C_1,\ldots,C_K\), by applying the \(K\)-means algorithm to the kernel mean embeddings \(\mu_{F_1},\ldots,\mu_{F_n}\) with respect to the metric \(\|\;\|_\scrH\) in the RKHS \(\scrH\) generated by \(k\):

\begin{enumerate}[label={\bf Step \arabic*.}, itemsep=3mm, parsep=0mm, align=left, labelwidth=*, leftmargin=-0.5 mm]
\item Initialize the representatives \(G_1,\ldots,G_K \in \calM(\calX)\) of the \(K\) clusters by choosing \(K\) different distributions randomly from the sample.
\item Assign each observation \(F_i\) to its nearest representative \(G_j\) with respect to the \(\scrH\)-MMD metric \eqref{Squared_MMD}. Denote by \(C_j = \{F_{i_{1}},\ldots,F_{i_{n_j}}\}\) the sample elements closest to \(G_j\) in this metric.
\item Compute the new centroid \(G_j\) of the cluster \(C_j\) as the sample mean \(\bar F_j\) of the distributions in \(C_j\).
\end{enumerate}
Steps 2 and 3 are iterated until there are no changes in the assignment of sample elements to clusters (that is, when the \(K\)-means procedure converges to a local optimum). The algorithm is carried out several times, so that the initial centroids in Step 1 are allowed to change. From the different classifications and local optima obtained throughout these repetitions, we finally keep the clusters leading to the lowest Within Clusters Sum of Squares:
\begin{equation} \label{WCSS}
\WCSS = \sum_{j=1}^K \sum_{i=1}^{n_j} \|\mu_{F_{ji}}-\mu_{\bar F_j}\|^2_{\scrH}.
\end{equation}

\subsection{Technical details} \label{Subsection.Technicalities}

\emph{Computing the RKHS distances}

In practice, to apply the kernel \(K\)-means algorithm for distributional observations detailed in Section~\ref{Subsection.Algorithm}, it is only necessary to compute the maximum mean discrepancies \(\MMD(\scrH,F_i,F_\ell)=\|\mu_{F_i}-\mu_{F_\ell}\|_\scrH\) between all the possible pairs \((F_i,F_\ell)\) of sampled distributions.
Further, by \eqref{Squared_MMD} it suffices to calculate the Gram matrix \(\bK=(\bK_{i\ell})_{i,\ell=1}^n\), where
\begin{equation} \label{InnerProduct_RKHS}
\bK_{i\ell} = \langle\mu_{F_i},\mu_{F_\ell}\rangle_\scrH = E_{\bX\sim F_i,\bY\sim F_\ell} k(\bX,\bY) = \int_{\calX\times\calX} k(\bx,\by) \, dF_i(\bx) \, dF_\ell(\by),
\end{equation}
for \(i,\ell=1,\ldots,n\) and where \(\bX\) and \(\bY\) are independent.
There can be two situations: one in which the sampled distributions \(F_i\) are completely available (this is the typical case in a Monte Carlo study, where we know the generative models; see Section~\ref{Subsection.Simulations.Uni}). Then, depending on the complexity of the expression of the functions \(k\), \(F_i\) and \(F_\ell\), the integral \eqref{InnerProduct_RKHS} can be computed exactly or at least numerically approximated. The second situation takes place when the only information about each \(F_i\) comes from its sample \(\bX_{i1},\ldots,\bX_{i N_i}\) (this is the standard when dealing with real data). In this case, we approximate \eqref{InnerProduct_RKHS} by the following unbiased estimator (see \cite{BOOK_KernelMeanEmbedDistrib})
\begin{equation} \label{Est_InnerProduct_RKHS}
\hat\bK_{i\ell} = \left\{
\begin{array}{ll}
\displaystyle \frac{1}{N_iN_\ell} \sum_{j=1}^{N_i} \sum_{l=1}^{N_\ell} k(\bX_{ij},\bX_{\ell l}) & \mbox{if \(i\neq\ell\)} \\
\displaystyle \frac{1}{N_i(N_i-1)} \sum_{j=1}^{N_i} \sum_{\substack{l=1\\l\neq j}}^{N_i} k(\bX_{ij},\bX_{il}) & \mbox{if \(i=\ell\)}.
\end{array} \right.
\end{equation}


\emph{Selecting the kernel}

It is well known (see \cite{Sriperumbudur_etal_09}, \cite{Gretton_et_al_12b}) that the choice of the kernel \(k\) is crucial for the performance of a kernelized procedure. This will be evident too in the simulations of Section~\ref{Section.Simulations} and the real data analysis of Section~\ref{Section.RealData}.

To gain insight into the most adequate choice of the kernel in the \(K\)-means procedure proposed in Section~\ref{Subsection.Algorithm}, we have evaluated the clustering algorithm with different kernels (\(\|\;\|\) stands for the Euclidean norm in \(\bbR^p\)):
\begin{list}{\textbullet}{\leftmargin=5mm \itemindent=0em}
\item The Gaussian kernel
\[
k(\bx,\bx') = \exp\left(-\frac{\|\bx-\bx'\|^2}{2\sigma^2}\right), \quad \mbox{where \(\sigma>0\),}
\]
and the Laplace kernel
\[
k(\bx,\bx') = \exp\left(-\frac{\|\bx-\bx'\|}{\sigma}\right),  \quad \mbox{where \(\sigma>0\),}
\]
are widely used in machine learning. They are both strictly positive definite, translation invariant and characteristic (that is, the mapping \(\mu:P \mapsto \mu_P\) is injective); see \cite{BOOK_KernelMeanEmbedDistrib}.
\item The modified Gaussian (MG) kernel
\[
k(\bx,\bx') = \exp\left(-\|\bx-\bx'\|^2/2\right) + \|\bx\|^{\alpha} \|\bx'\|^{\alpha} \quad \mbox{where \(\alpha\geq1\)},
\]
metrizes the \(\alpha\)-Wasserstein space on \(\bbR^p\), the set of probability distributions on \(\bbR^p\) with finite \(\alpha\) moment (see, e.g., \cite{ModesteDombry24}). When \(p=1\) the \(\alpha\)-Wasserstein distance \(W_\alpha\) between two probability distributions on \(\bbR\) can be computed as the \(L^\alpha\) distance between their corresponding quantile functions.
That is, given two probability distributions \(P\) and \(Q\) in \(\calM(\bbR)\) with distribution functions \(F_P\) and \(F_Q\) respectively,
\( W_\alpha(P,Q) = \|F_P^{-1}-F_Q^{-1}\|_\alpha\) (see \cite{PanaretosZemel19}), where \(F^{-1}(q)=\inf\{x:F(x)\geq q\}\), \(0<q<1\), and
\[
\|F_P-F_Q\|_\alpha^\alpha = \int_\bbR (F_P(t)-F_Q(t))^\alpha dt .
\]
As indicated in Section~\ref{Section.Introduction}, the 1- and 2-Wasserstein distances have been employed when clustering distributional data. Thus, the choice of a kernel metrizing the Wasserstein distance seems natural.
\item The energy kernel (see, e.g., \cite{ModesteDombry24}, \cite{Sejdinovic_etal_13})
\[
k(\bx,\bx') = \frac{1}{2} \left( \|\bx\|^{2\alpha} + \|\bx'\|^{2\alpha} - \|\bx-\bx'\|^{2\alpha} \right), \quad \mbox{where \(0<\alpha<1\),}
\]
is characteristic (see, e.g., \cite{SzekelyRizzo23}). With this kernel the RKHS distance \eqref{Squared_MMD} coincides with the energy distance.
Also,  for \(\alpha=0.5\) it holds that \(\|\mu_P-\mu_Q\|_{\scrH}^2 = 2\|F_P-F_Q\|_2^2\).
\end{list}

\emph{The choice of the tuning parameter}

All the kernels depend on one or more tuning parameters (\(\sigma\) in the Gaussian and Laplace cases, \(\alpha\) in the energy and MG kernels). There can be different strategies to select an ``optimal'' value of the parameter, although the simulations (Section~\ref{Section.Simulations}) indicate that the proposed kernel clustering procedure is fairly robust with respect to the choice of the tuning parameter.

For the Gaussian and the Laplace kernels, in this paper the tuning parameter has been chosen as the median, \(\sigma^*\), of the \(n\) standard deviations corresponding to the sample \(F_1,\ldots,F_n\) of probability distributions.
We have tried more elaborate selections of \(\sigma\).
For instance, we considered the partitions \(\calP_1,\ldots,\calP_M\) (derived from \(M\) different choices \(\sigma_1,\ldots,\sigma_M\) of \(\sigma\)) as a cluster ensemble to be combined into a final unique partition.
Specifically, for a grid of \(M=5\) parameter values, \(\sigma_m=\sigma^* \cdot 10^{m-3}, m=1,\ldots,5\), we used a clustering ensemble algorithm called the median partition approach with respect to the dissimilarity given by the symmetric difference of the partitions (see \cite{VegaPRuizS11}). Since the simulation results obtained with this strategy did not improve those obtained with \(\sigma^*\), we do not report them in this work.

For the energy kernel, we have selected three values of the tuning parameter for illustration purposes, \(\alpha=0.5\) (due to the equivalence between the resulting MMD and the \(L^2\) distance between the distribution functions), \(\alpha=0.25\) and \(\alpha=0.75\). Finally, for the MG kernel, the chosen values of the parameter are \(\alpha=2\) and \(\alpha=3\).

\emph{Evaluating the performance of the clustering algorithm}

When carrying out an unsupervised classification procedure, there are different ways to evaluate its behaviour. In the simulation studies of Section~\ref{Section.Simulations} and in the analysis of the SAR images (Section~\ref{Section.RealData}), we know the real group to which the observations belong. Thus, we need a measure of the similarity between two clusterings of the same sample. Here we have used the adjusted Rand index (ARI) of \cite{HubertArabie85} and the accuracy. The accuracy of a sample clustering with \(K\) clusters with respect to the original \(K\) groups is computed by mapping each cluster to a population group (there are \(K!\) possible mappings) and computing the total proportion of correct classifications. Of the \(K!\) available assignations, the one with highest such proportion is selected.


\section{Simulation studies} \label{Section.Simulations}

The aim of this section is to illustrate, via Monte Carlo experiments, the performance of our proposed kernel \(K\)-means algorithm for distributional data. The results give an intuition \textcolor{red}{insight?} on the effect of choosing a certain  kernel and its tuning parameter.

\subsection{Univariate distributions} \label{Subsection.Simulations.Uni}

First, we consider the case in which each realization of the random distribution \(F\) is defined on \(\bbR\) (i.e., \(p=1\)). For one-dimensional distributions we can compare the results of the clustering algorithm proposed in Section~\ref{Subsection.Algorithm} with those of the \(K\)-means procedure with respect to the Wasserstein metric.

The generative model employed in this section is, essentially, that given in \citet[Sect. 5.1]{GohVidal08} (see also \cite{ChenHung14}). The distributional data are \(n=100\) mixtures of uniform distributions whose support is contained in the interval [0,1000]. We denote by \(U(a,b)\) the uniform distribution on the interval \([a,b]\). The construction of each sampled distribution is described next. First, a global mixing parameter \(\lambda\in(0,1)\) is fixed. Next, for each replication \(j=1,\ldots,100\) of the Monte Carlo experiment, four random quantities  \(\lambda_{1j},\ldots,\lambda_{4j}\) are independently drawn from a \(U(0,4)\). These four \(\lambda\)'s are held fixed for the generation of the \(n=100\) mixtures in each replication. Then, the following two uniform densities are constructed, for \(i=1,\ldots,n/2=50\),
\begin{equation} \label{UniformsChenHung}
\begin{array}{llll}
f_{ij}^{(1)} \sim U(a_{ij},b_{ij}), & \mbox{where} & a_{ij} = 4(i-1)+\lambda_{1j}, &
b_{ij} =195+5i+\lambda_{2j}, \\ [2 mm]
f_{ij}^{(2)} \sim U(c_{ij},d_{ij}), & \mbox{where} & c_{ij} = 805-5i-\lambda_{3j}, &
d_{ij} =1004-4i-\lambda_{4j}.
\end{array}
\end{equation}
The final distributional dataset is given by
\begin{equation} \label{ClassesModelChenHung}
\begin{array}{lll}
\mbox{Class 1:} & g_{ij}^{(1)} = f_{ij}^{(1)} & i=1,\ldots,50, \\
\mbox{Class 2:} & g_{ij}^{(2)} = \lambda f_{ij}^{(1)} + (1-\lambda) f_{ij}^{(2)} & i=1,\ldots,50.
\end{array}
\end{equation}
Observe that, as \(\lambda\) grows from 0 to 1, cluster 2 merges into cluster 1.

\begin{figure}[h]
\begin{center}
\begin{tabular}{ccc}
\includegraphics[trim= 6 6 8 25,clip,width=0.3\textwidth]{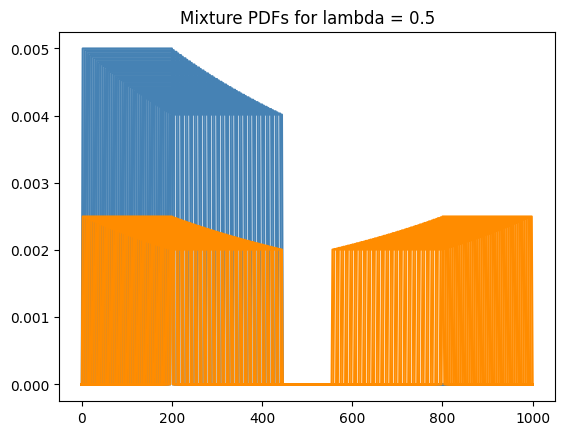} &
\includegraphics[trim= 6 6 8 25,clip,width=0.29\textwidth]{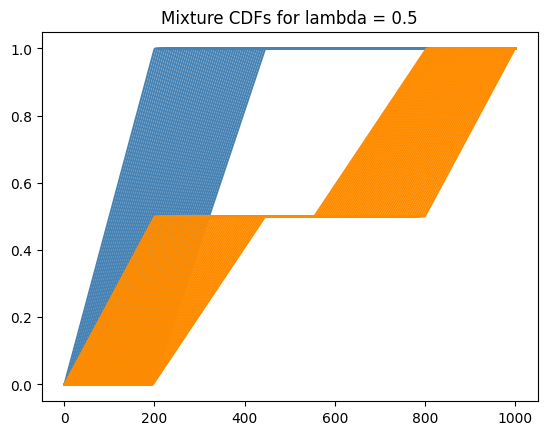} &
\includegraphics[trim= 6 6 8 25,clip,width=0.3\textwidth]{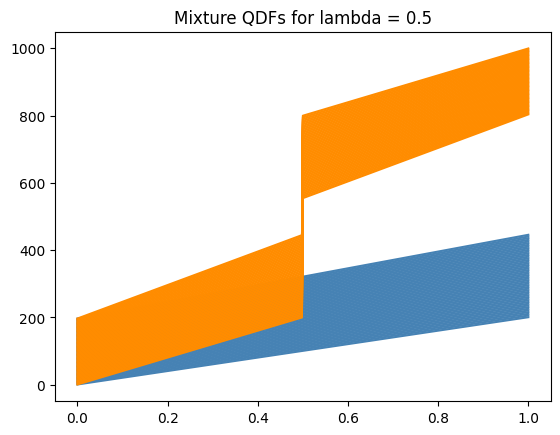} \\
(a) & (b) & (c)
\end{tabular}
\end{center}
\caption{(a) Densities, (b) cumulative distribution functions and (c) quantile functions sampled from the model \eqref{ClassesModelChenHung} with \(\lambda=0.5\). Class 1 (resp. 2) is in blue (resp. orange).}
\label{Figure.1dimModel.ChenHung}
\end{figure}

\begin{figure}[h]
\begin{center}
\begin{tabular}{ccc}
\includegraphics[trim= 6 6 8 28,clip,width=0.3\textwidth]{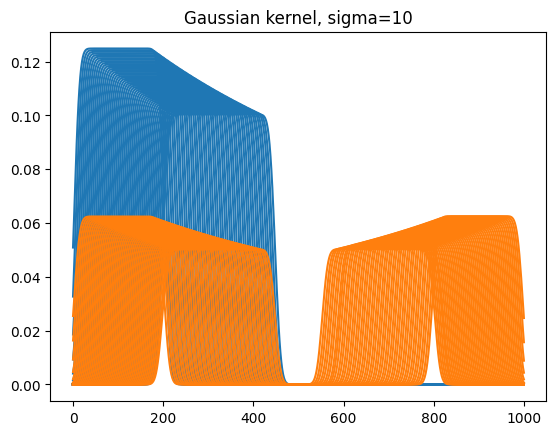} &
\includegraphics[trim= 6 6 8 29,clip,width=0.295\textwidth]{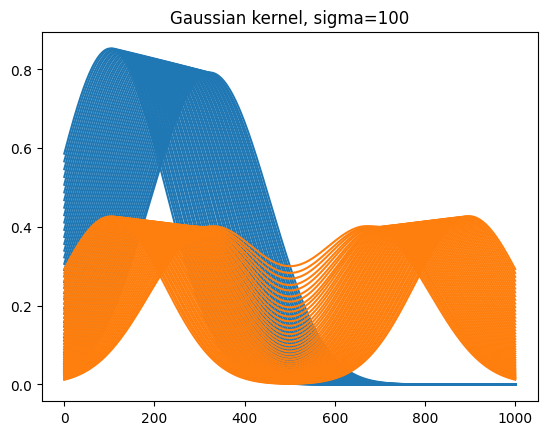} &
\includegraphics[trim= 6 6 8 28,clip,width=0.3\textwidth]{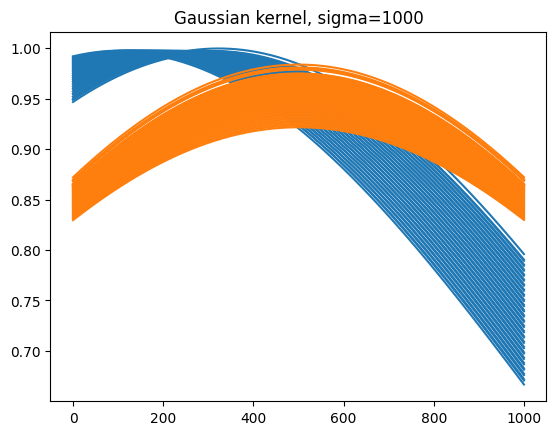} \\
(a) & (b) & (c)
\end{tabular}
\end{center}
\caption{KMEs of distributional data sampled from the model \eqref{ClassesModelChenHung} for \(\lambda=0.5\). Gaussian kernel with \(\sigma=\) (a) 10, (b) 100, (c) 1000. Class 1 (resp. 2) is in blue (resp. orange).}
\label{Figure.1dimKMEGaussian}
\end{figure}

Figure~\ref{Figure.1dimModel.ChenHung} displays one replication of \(n=100\) densities generated according to the model \eqref{ClassesModelChenHung} with \(\lambda=0.5\), as well as their distribution functions and their quantile functions. Figure~\ref{Figure.1dimKMEGaussian} shows the kernel mean embeddings of those distributional observations when using the Gaussian kernel with \(\sigma=\) 10, 100 and 1000. An analogous figure obtained with the Laplacian kernel can be found in the supplementary material and shows that, at least for model \eqref{ClassesModelChenHung}, the effect of the two kernels is similar. Further figures of the KMEs for various kernels in the cases \(\lambda=0.5\) and 0.9 have been included in the supplementary material.

We have carried out the following simulation study to check the performance of the kernel \(K\)-means algorithm,  with \(K=2\), when dealing with univariate distributional data.
For each of 100 Monte Carlo replications, a sample of size \(n=100\) was generated as in \eqref{UniformsChenHung} and \eqref{ClassesModelChenHung}, and classified according to the clustering algorithm of Section~\ref{Subsection.Algorithm}.
Each \(K\)-means procedure was initiated randomly 10 times, from which the centroids and clusters with lowest WCSS \eqref{WCSS} were selected.
In the Supplementary Material, for a single replication, there is a figure with the centroids obtained in the RKHS for different values of \(\lambda\) and \(\sigma\), the tuning parameter of the Gaussian kernel (see Section~\ref{Subsection.Technicalities}). Denoting \(\sigma_m=\sigma^* \cdot 10^{m-3}, m=1,\ldots,5\), the results are similar for values of \(\sigma\) between \(\sigma_1\) and \(\sigma_3\), but for \(\sigma_5\), the largest parameter, the kernel mean embeddings of the two classes have much larger variability, which might potentially spoil the classification.
In Table~\ref{tab:chenhung_coarse} we report the mean accuracy over all the replications. The metrics used in the clustering procedure are the 2-Wasserstein (2-W) metric and the MMD corresponding to the Gaussian, Laplace, modified Gaussian (MG) and energy kernels.

\begin{table}[h]
\begin{center}
\begin{tabular}{|c|c|c|c|c|c|c|c|c|}
\multicolumn{2}{c}{} & \multicolumn{1}{c}{Gaussian} & \multicolumn{1}{c}{Laplace} & \multicolumn{2}{c}{MG} & \multicolumn{3}{c}{Energy} \\
\multicolumn{1}{c}{$\lambda$} & \multicolumn{1}{c}{2-W} & \multicolumn{1}{c}{\(\sigma^*\)} & \multicolumn{1}{c}{\(\sigma^*\)} & \multicolumn{1}{c}{$\alpha=2$} & \multicolumn{1}{c}{$\alpha=3$} & \multicolumn{1}{c}{$\alpha=0.25$} & \multicolumn{1}{c}{$\alpha=0.5$} & \multicolumn{1}{c}{$\alpha=0.75$} \\ \hline
 0.0 & 1.0 & 1.0 & 1.0  & 1.0  & 1.0 & 1.0 & 1.0 & 1.0 \\
 0.1 & 1.0 & 1.0 & 1.0  & 1.0  & 1.0 & 1.0 & 1.0 & 1.0 \\
 0.2 & 1.0 & 1.0 & 1.0  & 1.0  & 1.0 & 1.0 & 1.0 & 1.0 \\
 0.3 & 1.0 & 1.0 & 1.0  & 1.0  & 1.0 & 1.0 & 1.0 & 1.0 \\
 0.4 & 1.0 & 1.0 & 1.0  & 1.0  & 1.0 & 1.0 & 1.0 & 1.0 \\
 0.5 & 1.0 & 1.0 & 1.0  & 1.0  & 1.0 & 1.0 & 1.0 & 1.0 \\
 0.6 & 1.0 & 1.0 & 0.5196 & 1.0  & 1.0 & 1.0 & 1.0 & 1.0 \\
 0.7 & 1.0 & 0.5067 & 0.51 & 1.0  & 1.0 & 0.5659 & 0.8617 & 0.9018 \\
 0.8 & 1.0 & 0.5013 & 0.5 & 1.0  & 1.0 & 0.5300 & 0.5800 & 0.7076 \\
 0.9 & 1.0 & 0.5 & 0.5 & 0.86 & 1.0 & 0.5100 & 0.5300 & 0.57 \\ \hline
\end{tabular}
\end{center}
\caption{Mean classification accuracy for the model in Section~\ref{Subsection.Simulations.Uni}.}
\label{tab:chenhung_coarse}
\end{table}

From the simulation results we conclude that, for this model, the \(K\)-means procedure performs best with the 2-Wasserstein metric and the MMD using the modified Gaussian kernel with \(\alpha=3\), closely followed by the MG kernel with \(\alpha=2\). In the Supplementary Material we have added the mean accuracy for values of \(\lambda\) between 0.9 and 0.99, employing the 2-Wasserstein and the MMD metrics with the MG kernel. The MG kernel with \(\alpha=3\) leads to the most stable results of the three procedures. The proportion of correct classifications with the Gaussian, Laplace and energy kernels is fairly similar: at around \(\lambda=0.7\) the mean accuracy drops sharply from 1 to 50\%. The energy kernel with \(\alpha=0.75\) is the best performing one of these five.

To conclude this section of univariate distributions, we report the results of an experiment with a modification of the model given by \eqref{UniformsChenHung} and \eqref{ClassesModelChenHung}.
The aim was to find cases where there were differences in the \(K\)-means performance between the 2-Wasserstein metric and the MMD.
The variations we considered are done by simply changing the starting values of $c_{ij}$ and $d_{ij}$ in \eqref{UniformsChenHung}.
In the original experiment, there are two fixed values, namely $C_0 = 805$ and $D_0 = 1004$.
The variations considered are (1) $C_0 = 100$ and $D_0 = 600$ and (2) $C_0 = 200$ and $D_0 = 600$.
In Table~\ref{tab:chenhung_var1} we present the results of the experiment for Variation 1. The results for Variation 2 are similar and appear in the Supplementary Material, as well as plots of one replication of \(n=100\) distributional data from Variation 1. What we observe in both tables is that there are values of \(\lambda\) (in the range 0--0.3 for Table~\ref{tab:chenhung_var1}) where \(K\)-means clustering with the 2-Wasserstein metric gives low correct classification rates ($\lessapprox$60\%), but it is outperformed by \(K\)-means with the MMD metric for the Laplace kernel. Unfortunately, for this model there is no kernel performing uniformly better than the others for all the values of \(\lambda\).

\begin{table}[h]
\centering
\begin{tabular}{|c|c|c|c|c|c|c|c|c|}
\multicolumn{2}{c}{} & \multicolumn{1}{c}{Gaussian} & \multicolumn{1}{c}{Laplace} & \multicolumn{2}{c}{MG} & \multicolumn{3}{c}{Energy} \\
\multicolumn{1}{c}{$\lambda$} & \multicolumn{1}{c}{2-W} & \multicolumn{1}{c}{\(\sigma^*\)} & \multicolumn{1}{c}{\(\sigma^*\)} & \multicolumn{1}{c}{$\alpha=2$} & \multicolumn{1}{c}{$\alpha=3$} & \multicolumn{1}{c}{$\alpha=0.25$} & \multicolumn{1}{c}{$\alpha=0.5$} & \multicolumn{1}{c}{$\alpha=0.75$} \\ \hline
0.0 & 0.5654 & 0.7700 & 0.7700 & 0.6324 & 0.6792 & 0.5632 & 0.5540 & 0.5497 \\
0.1 & 0.5045 & 0.7592 & 0.7501 & 0.6305 & 0.6688 & 0.7415 & 0.5637 & 0.5528 \\
0.2 & 0.5557 & 0.7400 & 0.7385 & 0.6310 & 0.6599 & 0.7311 & 0.7285 & 0.5653 \\
0.3 & 0.6008 & 0.5819 & 0.7208 & 0.7700 & 0.8147 & 0.7297 & 0.7202 & 0.7140 \\
0.4 & 0.6390 & 0.6274 & 0.5104 & 0.7600 & 0.8075 & 0.7196 & 0.7173 & 0.7100 \\
0.5 & 0.6734 & 0.5107 & 0.5110 & 0.7582 & 0.7992 & 0.5220 & 0.7100 & 0.7101 \\
0.6 & 0.7027 & 0.5000 & 0.5102 & 0.7500 & 0.7899 & 0.5179 & 0.5215 & 0.5419 \\
0.7 & 0.7295 & 0.5100 & 0.5100 & 0.5143 & 0.7798 & 0.5100 & 0.5105 & 0.5200 \\
0.8 & 0.7987 & 0.5000 & 0.5000 & 0.5100 & 0.5100 & 0.5100 & 0.5100 & 0.5100 \\
0.9 & 0.8400 & 0.5000 & 0.5000 & 0.5099 & 0.5057 & 0.5000 & 0.5006 & 0.5000 \\ \hline
\end{tabular}
\caption{Mean classification accuracy for Variation 1 of the model in Section~\ref{Subsection.Simulations.Uni}. }
\label{tab:chenhung_var1}
\end{table}

\subsection{Bivariate distributions} \label{Subsection.Simulations.Bi}


In this section we provide results from the kernel \(K\)-means algorithm (see Section~\ref{Subsection.Algorithm}) applied to synthetic distributions on \(\bbR^2\), generated by the model of Experiment 1 in \cite{Irpino_et_al_14}.
The simulation experiment consists of 100 Monte Carlo generations. Each of these replications is composed by 3 (known) clusters of 50 objects each, in sum, a total sample size of \(n=150\).

The first step of the Monte Carlo experiment is to choose, for each of the 3 groups and each of the 2 variables, four baseline parameters (mean, standard deviation, skewness and kurtosis) and the corresponding four standard deviations. These are used, in each Monte Carlo replication and for each cluster \(j=1,2,3\) and each variable \(X_\ell\), \(\ell=1,2\), to generate \(n/3=50\) collections of normally distributed parameters \((\mu_\ell^{(j)},\sigma_\ell^{(j)},\gamma_\ell^{(j)},\kappa_\ell^{(j)})\), which will be the mean, the standard deviation, the skewness and the kurtosis, respectively, of a Pearson distribution. In Table~\ref{tab:experiment1} we report the specific Gaussian distributions used to generate these parameters, which in \cite{Irpino_et_al_14} were independent random variables. Finally, a random vector \(\bX=(X_1,X_2)'\) with independent components is sampled, by drawing \(N_\ell=1000\) independent observations of each \(X_\ell\), \(\ell=1,2\), from a Pearson distribution with the parameters generated according to Table~\ref{tab:experiment1}. To summarize this simulation scheme, there are 100 Monte Carlo generations, each of which consists of \(n=150\) samples (in 3 groups of 50 each) of size 1000 from the bivariate vector \(\bX\). These samples are used to compute the estimator given in \eqref{Est_InnerProduct_RKHS}. Note the difference with respect to the data generated in Section~\ref{Subsection.Simulations.Uni}, where the distributional observations were fully known.

\begin{table}[h]
\centering
\begin{tabular}{|c|c|c|c|c|c|}
\hline
Cluster & Variable & Mean & Std. dev. & Skewness & Kurtosis \\
\(j\) & \(\ell\) & \(\mu_\ell^{(j)}\) & \(\sigma_\ell^{(j)}\) & \(\gamma_\ell^{(j)}\) & \(\kappa_\ell^{(j)}\) \\
\hline
\multirow{2}{*}{1} & 1 & \(N\)(-4.8, 6) & \(N\)(12, 1.2) & \(N\)(-0.05, 0.1) & \(N\)(3.10, 0.1) \\
\cline{2-6}
& 2 & \(N\)(17, 12) & \(N\)(6.0, 0.6) & \(N\)(0.00, 0.1) & \(N\)(2.95, 0.1) \\ \hline
\multirow{2}{*}{2} & 1 & \(N\)(-4.8, 6) & \(N\)(9, 1.2) & \(N\)(0.00, 0.1) & \(N\)(3.00, 0.1) \\
\cline{2-6}
& 2 & \(N\)(-17, 12) & \(N\)(4.6, 0.6) & \(N\)(0.00, 0.1) & \(N\)(3.00, 0.1) \\ \hline
\multirow{2}{*}{3} & 1 & \(N\)(10.0, 6) & \(N\)(6, 1.2) & \(N\)(0.10, 0.1) & \(N\)(2.95, 0.1) \\
\cline{2-6}
& 2 & \(N\)(0, 12) & \(N\)(3.3, 0.6) & \(N\)(-0.1, 0.1) & \(N\)(3.10, 0.1) \\ \hline
\end{tabular}
\caption{Independent univariate normal distributions generating the Pearson parameters of the synthetic data in Section~\ref{Subsection.Simulations.Bi}.}
\label{tab:experiment1}
\end{table}

The results of the Monte Carlo study are reported in Table~\ref{tab:irpino}. The number of random initializations in the kernel \(K\)-means procedure is 50. Contrary to the results in Section~\ref{Subsection.Simulations.Uni}, in this case the modified Gaussian kernel yields the worst results for both choices of its parameter \(\alpha\). The best results are attained by the energy kernel with \(\alpha=0.5\), but the results are comparable to those of the energy kernel with \(\alpha=0.25\) and \(\alpha=0.75\) and of the Gaussian and Laplacian kernels and slightly worse than the accuracy of 0.7867 obtained with the standard algorithm of \cite{Irpino_et_al_14}. We have also carried out the clustering procedures just with the samples of the first component \(X_1\) (a 1-dimensional setting similar to that of Section~\ref{Subsection.Simulations.Uni}). The results are summarized in Table~\ref{tab:irpino_1d} and are coherent with those of Table~\ref{tab:irpino} in the 2-dimensional setting: the performance of the modified Gaussian kernel is notably the worst and the rest of the kernel \(K\)-means procedures (as well as \(K\)-means based on the 2-Wasserstein metric) have an identical behaviour.

\begin{table}[h]
\centering
\begin{tabular}{|l|c|c|}
    \hline
    \multicolumn{1}{|c|}{Kernel} & Mean Accuracy & Mean Adjusted Rand Index \\
    \hline
    Gaussian \(\sigma^*\) & 0.7609 & 0.4381 \\
    Laplace \(\sigma^*\) & 0.7636 & 0.4447 \\
    MG $\alpha=2$ & 0.4566 & 0.0480 \\
    MG $\alpha=3$ & 0.4243 & 0.0248 \\
    Energy $\alpha=0.25$ & 0.7717 & 0.4520 \\
    Energy $\alpha=0.5$ & 0.7739 & 0.4511 \\
    Energy $\alpha=0.75$ & 0.7706 & 0.4434 \\
    \hline
\end{tabular}
\caption{Results of the simulations for the first model in Section~\ref{Subsection.Simulations.Bi}.}
\label{tab:irpino}
\end{table}

\begin{table}[h]
\centering
\begin{tabular}{|l|c|c|}
    \hline
    \multicolumn{1}{|c|}{Model}            & Mean Accuracy & Mean Adjusted Rand Index \\
    \hline
    2-W & 0.5895 & 0.2591 \\
    Gaussian \(\sigma^*\) & 0.5955 & 0.2821 \\
    Laplace \(\sigma^*\) & 0.5954 & 0.2818 \\
    MG $\alpha=2$ & 0.4165 & 0.0369 \\
    MG $\alpha=3$ & 0.4053 & 0.0246 \\
    Energy $\alpha=0.25$ & 0.5939 & 0.2738 \\
    Energy $\alpha=0.5$ & 0.5949 & 0.2686 \\
    Energy $\alpha=0.75$ & 0.5930 & 0.2645 \\
    \hline
\end{tabular}
\caption{Results of the simulations for the first component of the model in Section~\ref{Subsection.Simulations.Bi}.}
\label{tab:irpino_1d}
\end{table}

Note that in the model proposed in Experiment 1 of \cite{Irpino_et_al_14}, the one used up to now in this section, the two components \(X_1\) and \(X_2\) of the vector \(\bX\) are independent. The clustering algorithms considered by \cite{Irpino_et_al_14} depend critically on this assumption, as the criteria that these authors minimize are expressed in terms of a component-wise sum. Thus, the multivariate clustering methods employed in \cite{Irpino_et_al_14} are insensitive to the existence of dependence between the components of \(\bX\) and, consequently, to cluster information contained in the dependency structure of the vector. In contrast, our proposed kernel \(K\)-means algorithm takes into account such distributional information via the MMD distance given in \eqref{Squared_MMD}. Next, we illustrate the behaviour of the procedure for 2-dimensional distributional data with dependence, through simulations where the model of \cite{Irpino_et_al_14} has been modified to incorporate a non-zero correlation.

The sampling procedure in this variation is very similar to that of Experiment 1 of \cite{Irpino_et_al_14}, but with the following modifications to the generating model:

- There are just two bivariate populations, \(j=1,2\), (instead of 3) generating the samples of \(\bX\). The {\em only} difference between the two populations is the correlation between the two components of \(\bX\). The covariance matrices controlling this correlation are
\[
\bSigma_1 = \left( \begin{array}{rr}
1 & 0.9 \\ 0.9 & 1
\end{array} \right) \qquad \mbox{and} \qquad \bSigma_2 = \left( \begin{array}{rr}
1 & -0.9 \\ -0.9 & 1
\end{array} \right)
\]
for populations 1 and 2, respectively.
Apart from that, the distributions of the Pearson parameters are, to a large extent, given by the distribution of cluster 1 in Table~\ref{tab:experiment1} (see more information on these generating models below).

- The standard deviations of \(\mu_1^{(j)}\) and \(\mu_2^{(j)}\) are lower than those of \(\mu_1^{(1)}\) and \(\mu_2^{(2)}\) in Experiment 1 of \cite{Irpino_et_al_14}, so that the difference between Cluster 1 and 2, which is due to the correlation, is not masked by high variability.

The simulation experiment again consists of 100 Monte Carlo generations. Each of these replications is composed by \(n=100\) objects, 50 from each of the two populations. The object is, as before, a sample of size \(N_\ell=1000\) of the vector \(\bX\).
The procedure to generate one such sample from population \(j\), for \(j=1,2\), is as follows:
\begin{enumerate}
\item \label{Step1} We generate a normally distributed observation of each of the independent parameters \(\mu_\ell^{(j)}\), \(\sigma_\ell^{(j)}\), \(\gamma_\ell^{(j)}\) and \(\kappa_\ell^{(j)}\), \(\ell=1,2\). The specific Gaussian distributions appear in Table~\ref{tab:experiment_table}.
\item We sample $N_\ell=1000$ independent observations from $\bY_j=(Y_{j1},Y_{j2})'$, which follows a bivariate normal distribution N($\mbf{0}$,$\bSigma_j$).
\item We apply the cumulative distribution function \(\Phi\) of a standard normal to each observation of the coordinates \(Y_{j\ell}\), \(\ell=1,2\). Next, we evaluate the quantile function of a Pearson distribution (with the parameters obtained in Step~\ref{Step1}) to each observation of \(\Phi(Y_{j\ell})\). In this way, we obtain a sample of size \(N_\ell=1000\) of the vector \(\bX\), where the marginal distribution of the components is a Pearson distribution (as in Experiment 1 of \cite{Irpino_et_al_14}), but \(X_1\) and \(X_2\) have positive (resp. negative) correlation in population 1 (resp. 2).
\end{enumerate}

\begin{table}[H]
\centering
\begin{tabular}{|c|c|c|c|c|c|}
\hline
Cluster & Variable & Mean & Std. dev. & Skewness & Kurtosis \\
\(j\) & \(\ell\) & \(\mu_\ell^{(j)}\) & \(\sigma_\ell^{(j)}\) & \(\gamma_\ell^{(j)}\) & \(\kappa_\ell^{(j)}\) \\ \hline
\multirow{2}{*}{1} & 1 & N(-4.8, 0.5) & N(12, 1.2) & N(-0.05, 0.1) & N(3.10, 0.1) \\ \cline{2-6}
                   & 2 & N(17, 1)     & N(6, 0.6)  & N(0.00, 0.1)  & N(2.95, 0.1) \\ \hline
\multirow{2}{*}{2} & 1 & N(-4.8, 0.5) & N(12, 1.2) & N(-0.05, 0.1) & N(3.10, 0.1) \\ \cline{2-6}
                   & 2 & N(17, 1)     & N(6, 0.6)  & N(0.00, 0.1)  & N(2.95, 0.1) \\ \hline
    \end{tabular}
    \caption{Generating parameters for the variation of Experiment 1 in \cite{Irpino_et_al_14}.}
    \label{tab:experiment_table}
\end{table}

The results of the kernel \(K\)-means procedure for distributional data in this bidimensional example with dependence are displayed in Table~\ref{tab:pearsons_cluster1}. As it happened with the first model of this section, the performance of the modified Gaussian kernel is the worst. The Gaussian kernel with \(\sigma=\sigma^*\) and the energy one with \(\alpha=0.75\) also perform poorly. As a matter of fact, we checked that, as \(\alpha\) increases from 0.5 to 0.75, the accuracy of the \(K\)-means procedure with the energy kernel decreases. In the case of the Gaussian kernel, we also observed that with \(\sigma=0.1\sigma^*\), the accuracy increases with respect to that of \(\sigma=\sigma^*\). The best results are obtained with the energy kernel for \(\alpha=0.25\), followed by the energy kernel with \(\alpha=0.5\).

\begin{table}[H]
    \centering
    \begin{tabular}{|l|c|c|}
        \hline
        \multicolumn{1}{|c|}{Model}            & Mean Accuracy & Mean Adjusted Rand Index \\
        \hline
        Gaussian \(\sigma^*\) & 0.5678 & 0.0211 \\
        Laplace  \(\sigma^*\) & 0.8620 & 0.6559 \\
        MG $\alpha=2$         & 0.5395 & 0.0005 \\
        MG $\alpha=3$         & 0.5380 & -0.0001 \\
        Energy $\alpha=0.25$  & 0.9983 & 0.9932 \\
        Energy $\alpha=0.5$   & 0.9447 & 0.8585 \\
        Energy $\alpha=0.75$  & 0.5732 & 0.0250 \\
        \hline
    \end{tabular}
    \caption{Results of the simulations for the model with dependence in Section~\ref{Subsection.Simulations.Bi}.}
    \label{tab:pearsons_cluster1}
\end{table}


\section{Unsupervised classification of SAR images} \label{Section.RealData}

Synthetic aperture radar (SAR) is a remote sensing technology where a radar sends an eletromagnetic wave to the Earth surface and records afterwards the amount of energy reflected back, resulting in a grayscale image (see \cite{Moreira_etal_13}). The recorded measurement depends on the physical characteristics of the surface (sea, ice, mountain,\ldots) and its conditions (e.g., humidity). Some of the advantages of SAR images are their high resolution, the all-weather and day-and-night operation capability or their ability to penetrate surface layers.

Here we consider the problem of clustering a collection of SAR images using the information provided by distributional features derived from the image. The images are part of the TenGeoP-SARwv dataset (\cite{Wang_et_al_18}, \cite{Wang_et_al_19}), which consists of more than 37,000 SAR images from the Sentinel‐1A satellite of the European Copernicus program. Each TenGeoP-SARwv image was taken over the open ocean and labelled according to ten defined geophysical
phenomena, including both oceanic and meteorologic characteristics (see Table~\ref{tab:sar_dataset}).

\begin{table}[H]
    \centering
    \begin{tabular}{|c|c|c|}
        \hline
        \multicolumn{3}{|c|}{SAR Dataset}                 \\
        \hline
        Class & Description            & Number of images \\
        \hline
        F     & Pure Ocean Waves       & 4900             \\
        G     & Wind Streaks           & 4797             \\
        H     & Micro Convective Cells & 4598             \\
        I     & Rain Cells             & 4740             \\
        J     & Biological Slicks      & 4709             \\
        K     & Sea Ice                & 4370             \\
        L     & Iceberg                & 1980             \\
        M     & Low Wind Area          & 2160             \\
        N     & Atmospheric Front      & 4100             \\
        O     & Oceanic Front          & 1199             \\
        \hline
    \end{tabular}
    \caption{Classes and number of images available in the the TenGeoP-SARwv dataset.}
    \label{tab:sar_dataset}
\end{table}

Each of these SAR images consists of approximately $500 \times 500$ pixels.
Images from the TenGeoP-SARwv dataset can have different dimensions and are generally not square.
The shade of gray of any pixel is a value between $0$ and $65535$, quantifying the intensity, a measure of the energy that has returned to the sensor.
The probability distribution of the intensity per pixel is the first feature we use for the unsupervised classification of the SAR images (see, e.g., \cite{Li_et_al_16}, who address the problem of modeling this distribution in a SAR image).
In each image this gray level distribution will be approximated by the empirical distribution based on the sample constituted by the intensities of all the pixels in the image.
To reduce the computational load, we have discretized the intensity values as follows: first we have  multiplied each intensity by 256/65535 and then we have taken the integer part of the result as a first variable, \(X_1\), of interest.
The experiments in Section~\ref{SectionSAR_1} are carried out on cumulative distributions of this variable, that is, they are univariate as the simulations of Section~\ref{Subsection.Simulations.Uni}.

\subsection{Distributional samples from the gray level (\(X_1\))} \label{SectionSAR_1}

\begin{figure}[H]
    \centering
    \begin{tabular}{cc}
    \includegraphics[width=0.45\textwidth]{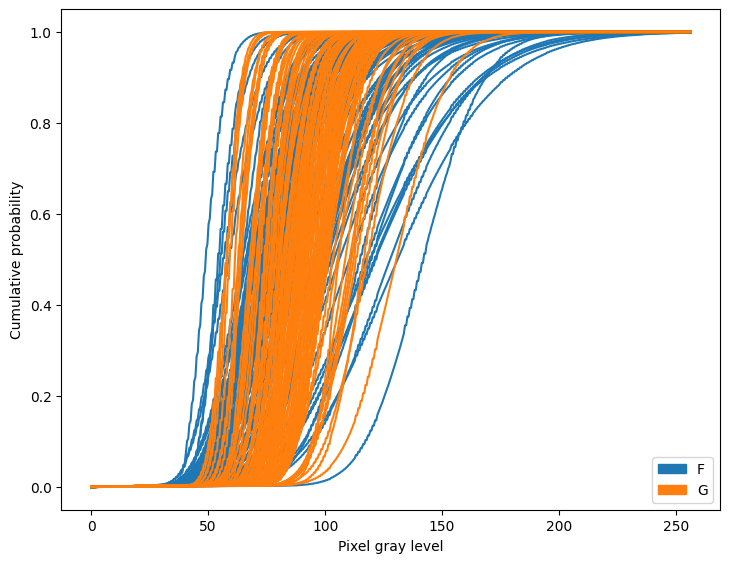} & \includegraphics[width=0.45\textwidth]{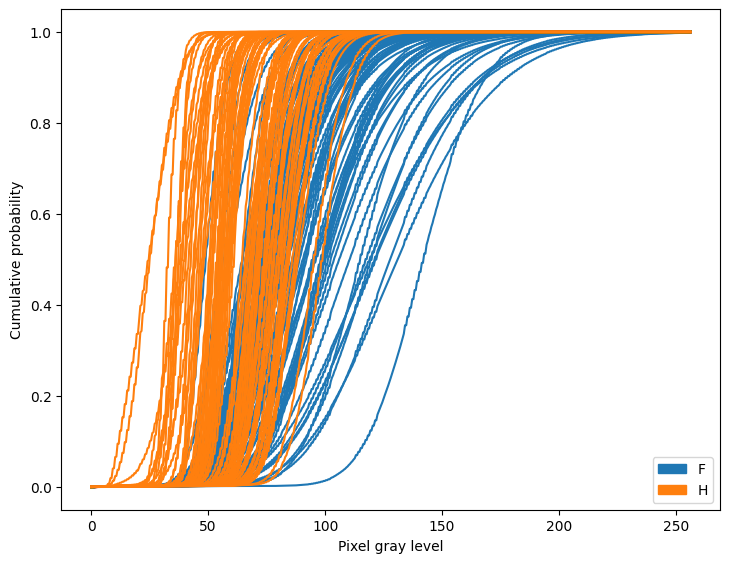} \\
    (a) Classes F and G & (b) Classes F and H
    \end{tabular}
    \caption{Samples of 100 distributions of from each of two classes in the TenGeoP-SARwv dataset.}
    \label{fig:cdf_classes_F_H}
\end{figure}

In this section the experiments check the performance of the proposed kernel \(K\)-means algorithm with subsets of two classes of the TenGeoP-SARwv dataset.
One of the classes is always F and the second one is any of the others appearing in Table~\ref{tab:sar_dataset}.
Then, for each generation, we randomly choose 100 images from each class. As an example, in Figure~\ref{fig:cdf_classes_F_H} we present the cdfs of the variable \(X_1\), the gray level per pixel, corresponding to 100 images from each of the classes F and G in (a) and F and H in (b).
We apply the \(K\)-means classification to each set of 200 images, setting $K=2$ for the number of clusters.
This is done with diverse metrics: the MMD with different kernels and the 2-Wasserstein.
In Table~\ref{tab:sar_unidimensional} we present the resulting mean accuracy for $100$ generations of the samples described above, for all the possible pairings of the class F with the other classes.
The \(K\)-means algorithm was run with 10 random initializations of the centroids.
Observe that the results derived with the modified Gaussian kernel are the worst. But the rest of the methods all have a comparable performance.

\begin{table}[h]
\centering
\begin{tabular}{|@{\hspace{1mm}}c@{\hspace{1mm}}|c|c|c|c|c|c|@{\hspace{1mm}}c@{\hspace{1mm}}|@{\hspace{1mm}}c@{\hspace{1mm}}|}
\hline
\multirow{3}{*}{Classes} & \multicolumn{8}{c|}{Mean Accuracy} \\ \cline{2-9}
 &  & Gaussian & Laplace & \multicolumn{2}{c|}{MG} & \multicolumn{3}{c|}{Energy} \\ \cline{5-6} \cline{7-9}
 & 2-W & $\sigma^*$ & $\sigma^*$ & $\alpha=2$ & $\alpha=3$ & $\alpha=0.25$ & $\alpha=0.5$ & $\alpha=0.75$ \\ \hline
\texttt{F, G} & 0.5340 & 0.5776 & 0.5790 & 0.5230 & 0.5230 & 0.5713 & 0.5615 & 0.5510 \\
\texttt{F, H} & 0.7546 & 0.7560 & 0.7544 & 0.6410 & 0.5952 & 0.7542 & 0.7499 & 0.7403 \\
\texttt{F, I} & 0.8236 & 0.8438 & 0.8411 & 0.6817 & 0.6026 & 0.8496 & 0.8461 & 0.8325 \\
\texttt{F, J} & 0.9725 & 0.9889 & 0.9896 & 0.9036 & 0.6586 & 0.9868 & 0.9844 & 0.9807 \\
\texttt{F, K} & 0.7893 & 0.7787 & 0.7775 & 0.6989 & 0.5933 & 0.7844 & 0.7877 & 0.7908 \\
\texttt{F, L} & 0.5951 & 0.6390 & 0.6383 & 0.5599 & 0.5480 & 0.6207 & 0.6077 & 0.5922 \\
\texttt{F, M} & 0.9906 & 0.9967 & 0.9968 & 0.9610 & 0.6681 & 0.9964 & 0.9952 & 0.9937 \\
\texttt{F, N} & 0.5520 & 0.5835 & 0.5873 & 0.5310 & 0.5280 & 0.5663 & 0.5555 & 0.5488 \\
\texttt{F, O} & 0.7702 & 0.7782 & 0.7659 & 0.6592 & 0.5982 & 0.8117 & 0.8097 & 0.7845 \\
\hline
\end{tabular}
\caption{Mean classification accuracy into \(K=2\) clusters of $100$ SAR images per class using only one variable, the intensity \(X_1\).}
\label{tab:sar_unidimensional}
\end{table}

\subsection{Distributional samples from the gray level (\(X_1\)) and its first derivative (\(X_2\))} \label{SectionSAR_2}

Here we consider the same SAR images as in Section~\ref{SectionSAR_1}, but now we add another variable per pixel, apart from its intensity value.
Specifically, we apply an edge detection filter to each image, which results in another image whose pixel values are obtained convolving the original image with the filter.
\cite{meester2022sar} indicate that there are multiple edge detection filters available.
A filter is defined by at least one matrix that is convolved with the image.
Generally, two matrices are used, one for the convolution in the horizontal direction and the other for the vertical direction.
In this work, we used the Sobel filter, which has the following two matrices for the vertical and horizontal directions respectively
\begin{equation} \label{Sobel_filter}
    \mbf{G}_x = \frac{1}{4}
    \left( \begin{array}{rrr}
        -1 & 0 & 1 \\
        -2 & 0 & 2 \\
        -1 & 0 & 1
    \end{array} \right), \qquad
    \mbf{G}_y = \frac{1}{4}
    \left( \begin{array}{rrr}
        -1 & -2 & -1 \\
        0  & 0  & 0  \\
        1  & 2  & 1
    \end{array} \right).
\end{equation}
The convolution with each of the matrices in \eqref{Sobel_filter} results in a gradient 2D vector for each pixel.
The euclidean norm, \(X_2\), of this gradient is then taken, resulting in the filtered image, a first-derivative edge detector.

To discretize the values of a filtered image to a new fixed interval, $[0, M]$ for some $M \in \bbN$, we proceed by considering the maximum theoretical value of applying the filter to the image.
Since the pixel value of an original SAR image takes values in $[0, 65535]$, then, one can manually check that a pixel neighbourhood configuration that attains the maximum value of this Sobel filter is, up to rotations and reflections,
\begin{equation*}
    \mbf{C}_{max} =
    \begin{bmatrix}
        0 & 0     & 65535 \\
        0 & \cdot & 65535 \\
        0 & 65535 & 65535
    \end{bmatrix}.
\end{equation*}
The resulting value of the filter for this configuration is
\begin{alignat*}{2}
    G_{max} & = \sqrt{(\mbf{G}_x \ast \mbf{C}_{max})^2 + (\mbf{G}_y \ast \mbf{C}_{max})^2}
    = \frac{65535 \sqrt{5}}{2} .
\end{alignat*}
Thus, the discretization of the filtered image is given by
\begin{equation}
    \text{discretized filter value} = \floor*{\frac{\text{original filter value}}{M/\left( 65535 \sqrt{5}/2 \right)}}.
\end{equation}
This method ensures that all filtered images are equally scaled.
Note it is not necessary that the discretization intervals for the filter and the pixel values coincide.

Next, since the distributional data now correspond to the bidimensional distribution of \((X_1,X_2)\), to ensure a feasible computation time, we consider rather small discretization intervals for \(X_1\) and \(X_2\).
We have observed that the filter values \((X_2)\) are rarely distributed near the maximum possible value.
This is logical, as a large value of the filter requires a very abrupt change in intensity. Therefore, to compensate for this, the discretization of the Sobel derivative is done in a larger range, [0,200], than that of the intensity, which we chose to be [0,100] for this bidimensional experiment. Otherwise, the discretized values of \(X_2\) would get too concentrated in the lower part of the range.

Table~\ref{tab:sar_bidimensional} displays the results of the clustering experiment described in Section~\ref{SectionSAR_1} using the bidimensional information. Comparing the resulting mean accuracy reported in Tables~\ref{tab:sar_unidimensional} and~\ref{tab:sar_bidimensional}, we observe that in the majority of cases this accuracy increases when employing the information provided by the derivative \(X_2\). In the cases when the accuracy does not increase with this second variable (mainly corresponding to the pairs ({\texttt F},{\texttt J}) and ({\texttt F},{\texttt M})), the relative variation experienced by the accuracy is always below 1.7\%. Thus, again we see that the kernel \(K\)-means procedure for distributional data can be applied for multidimensional distributions, thus potentially increasing the performance of the unsupervised classification in the one-dimensional case.

\begin{table}[h]
\centering
\begin{tabular}{|@{\hspace{1mm}}c@{\hspace{1mm}}|c|c|c|c|c|@{\hspace{1mm}}c@{\hspace{1mm}}|@{\hspace{1mm}}c@{\hspace{1mm}}|}
\hline
\multirow{3}{*}{Classes} & \multicolumn{7}{c|}{Mean Accuracy} \\ \cline{2-8}
 & Gaussian & Laplace & \multicolumn{2}{c|}{MG} & \multicolumn{3}{c|}{Energy} \\ \cline{4-5} \cline{6-8}
 & $\sigma^*$ & $\sigma^*$ & $\alpha=2$ & $\alpha=3$ & $\alpha=0.25$ & $\alpha=0.5$ & $\alpha=0.75$ \\ \hline
\texttt{F, G} & 0.7181 & 0.7331 & 0.5333 & 0.5320 & 0.7366 & 0.7283 & 0.7195 \\
\texttt{F, H} & 0.8727 & 0.8360 & 0.6672 & 0.5952 & 0.8394 & 0.8610 & 0.8740 \\
\texttt{F, I} & 0.9205 & 0.9184 & 0.7032 & 0.5998 & 0.9177 & 0.9220 & 0.9199 \\
\texttt{F, J} & 0.9789 & 0.9862 & 0.8977 & 0.6512 & 0.9853 & 0.9822 & 0.9769 \\
\texttt{F, K} & 0.8299 & 0.8100 & 0.7099 & 0.5956 & 0.8045 & 0.8154 & 0.8261 \\
\texttt{F, L} & 0.7348 & 0.7426 & 0.5808 & 0.5540 & 0.7440 & 0.7439 & 0.7373 \\
\texttt{F, M} & 0.9923 & 0.9967 & 0.9500 & 0.6571 & 0.9960 & 0.9931 & 0.9913 \\
\texttt{F, N} & 0.6873 & 0.7068 & 0.5499 & 0.5362 & 0.7069 & 0.7000 & 0.6922 \\
\texttt{F, O} & 0.8736 & 0.8919 & 0.6716 & 0.5954 & 0.8951 & 0.8883 & 0.8726 \\
\hline
\end{tabular}
\caption{Mean classification accuracy into \(K=2\) clusters of $100$ SAR images per class using the intensity \(X_1\) and the first derivative \(X_2\).}
\label{tab:sar_bidimensional}
\end{table}

\subsection{Selecting the number of clusters: a quick trial} \label{SectionSAR_K}

To conclude this work we focus on the problem of selecting the number of clusters when classifying real distributional data, where the correct partition is not available.
In this case we need an internal validation index, which generally compares the intra-cluster dispersion and the separation between the partitioned observations (see \cite{Arbelaitz_et_al_13}).
This matter has been studied extensively in the context of multivariate data, where diverse cluster validity indices have been proposed.
Naturally, the references on selecting the number of clusters when the data have infinite dimension are significantly more scarce (see, e.g., \cite{JacquesPreda14}, \cite{Zambom_et_al_22} and references therein for procedures in the functional data setting).

{\em The cluster validity indices}

For this computational experiment we have selected three internal cluster validity indices that performed well in the extensive study of \cite{Arbelaitz_et_al_13} and that are feasible to compute even for samples of multidimensional distributions: the Caliński-Harabasz, the silhouette and the Davies-Bouldin* indices. Their definitions are given below. We use the notation \(\scrC\) for the cluster partition \(\{C_1,\ldots,C_K\}\) of the sample \(F_1,\ldots,F_n\).
The notation \(\bar F = \sum_{i=1}^n F_i/n\) stands for the whole sample mean and \(d\) is the distance between the distributions.

\begin{list}{\textbullet}{\leftmargin=5mm \itemindent=0em}
\item \cite{CalinskiHarabasz74} index:
    \begin{equation}
    \mbox{CH}(\scrC) = \frac{B(\scrC)}{W(\scrC)} \cdot \frac{n-K}{K-1},
    \end{equation}
    where
    \(
    B(\scrC) = \sum_{j=1}^K n_j d(\bar F_j, \bar F)^2
    \)
    is the between-clusters sum of squares and
    \(
    W(\scrC) = \sum_{j=1}^K \sum_{i=1}^{n_j} d(F_{ji}, \bar F_j)^2
    \)
    is the within-clusters sum of squares.

\item Silhouette index (\cite{Rousseeuw87}):
    \begin{equation}
    \mbox{Sil}(\scrC)
    = \frac{1}{n} \sum_{j=1}^K \sum_{i=1}^{n_j} \frac{b(F_{ji}, C_j) - a(F_{ji}, C_j)}{\max\{a(F_{ji}, C_j), b(F_{ji}, C_j)\}},
    \end{equation}
    where
    \[
    a(F_{ji}, C_j) = \frac{1}{n_j - 1} \sum_{\substack{\ell=1\\\ell\neq i}}^j d(F_{ji}, F_{j\ell})
    \]
    is the mean distance of the distributional datum \(F_{ji}\) to the rest of the distributions in its cluster \(C_j\) and
    \[
    b(F_{ji}, C_j) = \min_{\substack{\ell=1,\ldots,K\\ \ell\neq j}} \left\{ \frac{1}{n_\ell} \sum_{m=1}^{n_\ell} d(F_{ji}, F_{\ell m}) \right\}.
    \]
    \item Davies-Bouldin* index (\cite{KimRamakrishna05}):
    \begin{equation}
        \mbox{DB*}(\scrC) = \frac{1}{K} \sum_{j=1}^K
        \frac{ \max \{ S(C_j) + S(C_\ell) : \ell=1,\ldots,K, \ell\neq j\} }
        { \max \{ d(\bar F_j, \bar F_\ell) : \ell=1,\ldots,K, \ell\neq j \} }
    \end{equation}
    where
    \(
    S(C_j) = \sum_{i=1}^{n_j} d(F_{ji}, \bar F_j) / n_j
    \) is the mean distance of the data in cluster \(C_j\) to its centroid \(G_j\).
\end{list}
In terms of these indices the best partition is given by the highest values of CH and Sil and the lowest one of DB$^*$.

{\em Experimental results}

Initially, the data selection procedure is the same as in the first experiment of Section~\ref{SectionSAR_1}. The gray-level distribution is determined in two classes of SAR images (one of them is always class F) after drawing randomly 100 data from each class.
Further, we also perform the experiment for two cases in which 100 images are sampled from three classes, \{F,G,H\} and \{F,J,M\}.
In this section the aim is to compare the performance of the three cluster validity indices for \(K\) = 2, 3 and 4. The proportion of times that \(K\) = 2, 3 and 4 were chosen as the optimal number of clusters with the Caliński-Harabasz, silhouette and Davies-Bouldin* indices appear in Tables~\ref{tab:sar_unidimensional_calinski_harabasz_proportions}, \ref{tab:sar_unidimensional_silhouette_proportions} and~\ref{tab:sar_unidimensional_davies_bouldin_star_proportions} respectively, where the highest proportion is marked in bold. Clearly, the silhouette index outperforms by far the other two indices in selecting the correct number of groups.

\begin{table}[H]
\begin{tabular}{|c|r|r|r|r|r|r|r|r|r|}
\multicolumn{2}{@{}c@{}}{} & \multicolumn{1}{c}{Gaussian} & \multicolumn{1}{c}{Laplace} & \multicolumn{2}{@{}c@{}}{MG} & \multicolumn{3}{@{}c@{}}{Energy} \\
\multicolumn{1}{@{}c@{}}{Classes} & \multicolumn{1}{@{}c@{}}{2-W} & \multicolumn{1}{@{}c@{}}{\(\sigma^*\)} & \multicolumn{1}{@{}c@{}}{\(\sigma^*\)} & \multicolumn{1}{c}{$\alpha=2$} & \multicolumn{1}{c}{$\alpha=3$} & \multicolumn{1}{c}{$\alpha=0.25$} & \multicolumn{1}{c}{$\alpha=0.5$} & \multicolumn{1}{c}{$\alpha=0.75$} \\ \hline
F,G & $\begin{array}{c} 0.21 \\ 0.26 \\ \mbf{0.53} \end{array}$ & $\begin{array}{c} 0.14 \\ 0.30 \\ \mbf{0.56} \end{array}$ & $\begin{array}{c} 0.21 \\ 0.35 \\ \mbf{0.44} \end{array}$ & $\begin{array}{c} 0.00 \\ 0.00 \\ \mbf{1.00} \end{array}$ & $\begin{array}{c} 0.00 \\ 0.13 \\ \mbf{0.87} \end{array}$ & $\begin{array}{c} 0.08 \\ 0.26 \\ \mbf{0.66} \end{array}$ & $\begin{array}{c} 0.01 \\ 0.10 \\ \mbf{0.89} \end{array}$ & $\begin{array}{c} 0.00 \\ 0.00 \\ \mbf{1.00} \end{array}$ \\ \hline
F,H & $\begin{array}{c} 0.00 \\ 0.19 \\ \mbf{0.81} \end{array}$ & $\begin{array}{c} 0.09 \\ 0.02 \\ \mbf{0.89} \end{array}$ & $\begin{array}{c} 0.13 \\ 0.01 \\ \mbf{0.86} \end{array}$ & $\begin{array}{c} 0.00 \\ 0.00 \\ \mbf{1.00} \end{array}$ & $\begin{array}{c} 0.00 \\ 0.02 \\ \mbf{0.98} \end{array}$ & $\begin{array}{c} 0.04 \\ 0.06 \\ \mbf{0.90} \end{array}$ & $\begin{array}{c} 0.00 \\ 0.08 \\ \mbf{0.92} \end{array}$ & $\begin{array}{c} 0.00 \\ 0.05 \\ \mbf{0.95} \end{array}$ \\ \hline
F,I & $\begin{array}{c} 0.00 \\ 0.17 \\ \mbf{0.83} \end{array}$ & $\begin{array}{c} 0.21 \\ 0.10 \\ \mbf{0.69} \end{array}$ & $\begin{array}{c} 0.34 \\ 0.10 \\ \mbf{0.56} \end{array}$ & $\begin{array}{c} 0.00 \\ 0.01 \\ \mbf{0.99} \end{array}$ & $\begin{array}{c} 0.00 \\ 0.11 \\ \mbf{0.89} \end{array}$ & $\begin{array}{c} 0.10 \\ 0.10 \\ \mbf{0.80} \end{array}$ & $\begin{array}{c} 0.01 \\ 0.08 \\ \mbf{0.91} \end{array}$ & $\begin{array}{c} 0.00 \\ 0.06 \\ \mbf{0.94} \end{array}$ \\ \hline
F,J & $\begin{array}{c} 0.01 \\ \mbf{0.54} \\ 0.45 \end{array}$ & $\begin{array}{c} \mbf{1.00} \\ 0.00 \\ 0.00 \end{array}$ & $\begin{array}{c} \mbf{1.00} \\ 0.00 \\ 0.00 \end{array}$ & $\begin{array}{c} 0.00 \\ 0.00 \\ \mbf{1.00} \end{array}$ & $\begin{array}{c} 0.00 \\ 0.02 \\ \mbf{0.98} \end{array}$ & $\begin{array}{c} \mbf{0.98} \\ 0.02 \\ 0.00 \end{array}$ & $\begin{array}{c} 0.39 \\ \mbf{0.51} \\ 0.10 \end{array}$ & $\begin{array}{c} 0.02 \\ \mbf{0.68} \\ 0.30 \end{array}$ \\ \hline
F,K & $\begin{array}{c} 0.00 \\ 0.13 \\ \mbf{0.87} \end{array}$ & $\begin{array}{c} 0.09 \\ 0.02 \\ \mbf{0.89} \end{array}$ & $\begin{array}{c} 0.26 \\ 0.02 \\ \mbf{0.72} \end{array}$ & $\begin{array}{c} 0.00 \\ 0.00 \\ \mbf{1.00} \end{array}$ & $\begin{array}{c} 0.00 \\ 0.01 \\ \mbf{0.99} \end{array}$ & $\begin{array}{c} 0.09 \\ 0.07 \\ \mbf{0.84} \end{array}$ & $\begin{array}{c} 0.00 \\ 0.04 \\ \mbf{0.96} \end{array}$ & $\begin{array}{c} 0.00 \\ 0.01 \\ \mbf{0.99} \end{array}$ \\ \hline
F,L & $\begin{array}{c} 0.00 \\ 0.06 \\ \mbf{0.94} \end{array}$ & $\begin{array}{c} 0.00 \\ 0.05 \\ \mbf{0.95} \end{array}$ & $\begin{array}{c} 0.00 \\ 0.08 \\ \mbf{0.92} \end{array}$ & $\begin{array}{c} 0.00 \\ 0.01 \\ \mbf{0.99} \end{array}$ & $\begin{array}{c} 0.02 \\ 0.06 \\ \mbf{0.92} \end{array}$ & $\begin{array}{c} 0.00 \\ 0.04 \\ \mbf{0.96} \end{array}$ & $\begin{array}{c} 0.00 \\ 0.02 \\ \mbf{0.98} \end{array}$ & $\begin{array}{c} 0.00 \\ 0.01 \\ \mbf{0.99} \end{array}$ \\ \hline
F,M & $\begin{array}{c} 0.14 \\ 0.35 \\ \mbf{0.51} \end{array}$ & $\begin{array}{c} 0.00 \\ 0.00 \\ \mbf{1.00} \end{array}$ & $\begin{array}{c} 0.01 \\ 0.06 \\ \mbf{0.93} \end{array}$ & $\begin{array}{c} 0.00 \\ 0.01 \\ \mbf{0.99} \end{array}$ & $\begin{array}{c} 0.00 \\ 0.01 \\ \mbf{0.99} \end{array}$ & $\begin{array}{c} 0.12 \\ 0.00 \\ \mbf{0.88} \end{array}$ & $\begin{array}{c} 0.08 \\ 0.00 \\ \mbf{0.92} \end{array}$ & $\begin{array}{c} 0.07 \\ 0.01 \\ \mbf{0.92} \end{array}$ \\ \hline
F,N & $\begin{array}{c} 0.00 \\ 0.11 \\ \mbf{0.89} \end{array}$ & $\begin{array}{c} 0.00 \\ 0.13 \\ \mbf{0.87} \end{array}$ & $\begin{array}{c} 0.00 \\ 0.19 \\ \mbf{0.81} \end{array}$ & $\begin{array}{c} 0.00 \\ 0.00 \\ \mbf{1.00} \end{array}$ & $\begin{array}{c} 0.00 \\ 0.05 \\ \mbf{0.95} \end{array}$ & $\begin{array}{c} 0.00 \\ 0.14 \\ \mbf{0.86} \end{array}$ & $\begin{array}{c} 0.00 \\ 0.07 \\ \mbf{0.93} \end{array}$ & $\begin{array}{c} 0.00 \\ 0.01 \\ \mbf{0.99} \end{array}$ \\ \hline
F,O & $\begin{array}{c} 0.00 \\ 0.00 \\ \mbf{1.00} \end{array}$ & $\begin{array}{c} 0.00 \\ 0.12 \\ \mbf{0.88} \end{array}$ & $\begin{array}{c} 0.00 \\ 0.25 \\ \mbf{0.75} \end{array}$ & $\begin{array}{c} 0.00 \\ 0.00 \\ \mbf{1.00} \end{array}$ & $\begin{array}{c} 0.00 \\ 0.07 \\ \mbf{0.93} \end{array}$ & $\begin{array}{c} 0.00 \\ 0.02 \\ \mbf{0.98} \end{array}$ & $\begin{array}{c} 0.00 \\ 0.00 \\ \mbf{1.00} \end{array}$ & $\begin{array}{c} 0.00 \\ 0.00 \\ \mbf{1.00} \end{array}$ \\ \hline
F,G,H & $\begin{array}{c} 0.00 \\ 0.22 \\ \mbf{0.78} \end{array}$ & $\begin{array}{c} 0.01 \\ 0.18 \\ \mbf{0.81} \end{array}$ & $\begin{array}{c} 0.03 \\ 0.21 \\ \mbf{0.76} \end{array}$ & $\begin{array}{c} 0.00 \\ 0.00 \\ \mbf{1.00} \end{array}$ & $\begin{array}{c} 0.00 \\ 0.01 \\ \mbf{0.99} \end{array}$ & $\begin{array}{c} 0.01 \\ 0.23 \\ \mbf{0.76} \end{array}$ & $\begin{array}{c} 0.00 \\ 0.07 \\ \mbf{0.93} \end{array}$ & $\begin{array}{c} 0.00 \\ 0.00 \\ \mbf{1.00} \end{array}$ \\ \hline
F,J,M & $\begin{array}{c} 0.00 \\ 0.01 \\ \mbf{0.99} \end{array}$ & $\begin{array}{c} 0.00 \\ \mbf{0.54} \\ 0.46 \end{array}$ & $\begin{array}{c} 0.00 \\ \mbf{0.55} \\ 0.45 \end{array}$ & $\begin{array}{c} 0.00 \\ 0.35 \\ \mbf{0.65} \end{array}$ & $\begin{array}{c} 0.00 \\ 0.04 \\ \mbf{0.96} \end{array}$ & $\begin{array}{c} 0.00 \\ \mbf{0.76} \\ 0.24 \end{array}$ & $\begin{array}{c} 0.00 \\ 0.08 \\ \mbf{0.92} \end{array}$ & $\begin{array}{c} 0.00 \\ 0.02 \\ \mbf{0.98} \end{array}$ \\ \hline
\end{tabular}
\caption{Proportion of times each $K \in \{2,3,4\}$ (ordered from up $(K=2)$ to down $(K=4)$) scored best  (higher) in terms of the Caliński-Harabasz index for the models over $100$ generations of the SAR dataset with $100$ images per class.}
\label{tab:sar_unidimensional_calinski_harabasz_proportions}
\end{table}

\begin{table}[H]
\begin{tabular}{|c|r|r|r|r|r|r|r|r|r|}
\multicolumn{2}{@{}c@{}}{} & \multicolumn{1}{c}{Gaussian} & \multicolumn{1}{c}{Laplace} & \multicolumn{2}{@{}c@{}}{MG} & \multicolumn{3}{@{}c@{}}{Energy} \\
\multicolumn{1}{@{}c@{}}{Classes} & \multicolumn{1}{@{}c@{}}{2-W} & \multicolumn{1}{@{}c@{}}{\(\sigma^*\)} & \multicolumn{1}{@{}c@{}}{\(\sigma^*\)} & \multicolumn{1}{c}{$\alpha=2$} & \multicolumn{1}{c}{$\alpha=3$} & \multicolumn{1}{c}{$\alpha=0.25$} & \multicolumn{1}{c}{$\alpha=0.5$} & \multicolumn{1}{c}{$\alpha=0.75$} \\ \hline
F,G & $\begin{array}{c} \mbf{0.99} \\ 0.01 \\ 0.00 \end{array}$ & $\begin{array}{c} \mbf{0.70} \\ 0.21 \\ 0.09 \end{array}$ & $\begin{array}{c} \mbf{0.69} \\ 0.21 \\ 0.10 \end{array}$ & $\begin{array}{c} \mbf{0.97} \\ 0.02 \\ 0.01 \end{array}$ & $\begin{array}{c} \mbf{0.99} \\ 0.01 \\ 0.00 \end{array}$ & $\begin{array}{c} \mbf{0.85} \\ 0.13 \\ 0.02 \end{array}$ & $\begin{array}{c} \mbf{0.89} \\ 0.10 \\ 0.01 \end{array}$ & $\begin{array}{c} \mbf{0.94} \\ 0.06 \\ 0.00 \end{array}$ \\ \hline
F,H & $\begin{array}{c} 0.19 \\ \mbf{0.80} \\ 0.01 \end{array}$ & $\begin{array}{c} \mbf{0.89} \\ 0.08 \\ 0.03 \end{array}$ & $\begin{array}{c} \mbf{0.86} \\ 0.12 \\ 0.02 \end{array}$ & $\begin{array}{c} \mbf{0.66} \\ 0.19 \\ 0.15 \end{array}$ & $\begin{array}{c} \mbf{0.99} \\ 0.00 \\ 0.01 \end{array}$ & $\begin{array}{c} \mbf{0.76} \\ 0.14 \\ 0.10 \end{array}$ & $\begin{array}{c} \mbf{0.51} \\ 0.42 \\ 0.07 \end{array}$ & $\begin{array}{c} 0.22 \\ \mbf{0.75} \\ 0.03 \end{array}$ \\ \hline
F,I & $\begin{array}{c} 0.42 \\ \mbf{0.58} \\ 0.00 \end{array}$ & $\begin{array}{c} \mbf{0.96} \\ 0.02 \\ 0.02 \end{array}$ & $\begin{array}{c} \mbf{0.96} \\ 0.01 \\ 0.03 \end{array}$ & $\begin{array}{c} \mbf{0.64} \\ 0.26 \\ 0.10 \end{array}$ & $\begin{array}{c} \mbf{0.98} \\ 0.02 \\ 0.00 \end{array}$ & $\begin{array}{c} \mbf{0.78} \\ 0.21 \\ 0.01 \end{array}$ & $\begin{array}{c} 0.48 \\ \mbf{0.52} \\ 0.00 \end{array}$ & $\begin{array}{c} 0.30 \\ \mbf{0.67} \\ 0.03 \end{array}$ \\ \hline
F,J & $\begin{array}{c} \mbf{0.77} \\ 0.23 \\ 0.00 \end{array}$ & $\begin{array}{c} \mbf{1.00} \\ 0.00 \\ 0.00 \end{array}$ & $\begin{array}{c} \mbf{1.00} \\ 0.00 \\ 0.00 \end{array}$ & $\begin{array}{c} 0.01 \\ \mbf{0.66} \\ 0.33 \end{array}$ & $\begin{array}{c} 0.25 \\ 0.06 \\ \mbf{0.69} \end{array}$ & $\begin{array}{c} \mbf{1.00} \\ 0.00 \\ 0.00 \end{array}$ & $\begin{array}{c} \mbf{0.97} \\ 0.03 \\ 0.00 \end{array}$ & $\begin{array}{c} \mbf{0.90} \\ 0.10 \\ 0.00 \end{array}$ \\ \hline
F,K & $\begin{array}{c} \mbf{0.64} \\ 0.36 \\ 0.00 \end{array}$ & $\begin{array}{c} \mbf{1.00} \\ 0.00 \\ 0.00 \end{array}$ & $\begin{array}{c} \mbf{1.00} \\ 0.00 \\ 0.00 \end{array}$ & $\begin{array}{c} 0.12 \\ \mbf{0.52} \\ 0.36 \end{array}$ & $\begin{array}{c} \mbf{0.91} \\ 0.02 \\ 0.07 \end{array}$ & $\begin{array}{c} \mbf{1.00} \\ 0.00 \\ 0.00 \end{array}$ & $\begin{array}{c} \mbf{0.97} \\ 0.02 \\ 0.01 \end{array}$ & $\begin{array}{c} \mbf{0.79} \\ 0.20 \\ 0.01 \end{array}$ \\ \hline
F,L & $\begin{array}{c} \mbf{0.85} \\ 0.13 \\ 0.02 \end{array}$ & $\begin{array}{c} 0.15 \\ \mbf{0.56} \\ 0.29 \end{array}$ & $\begin{array}{c} 0.24 \\ \mbf{0.49} \\ 0.27 \end{array}$ & $\begin{array}{c} \mbf{0.98} \\ 0.01 \\ 0.01 \end{array}$ & $\begin{array}{c} \mbf{1.00} \\ 0.00 \\ 0.00 \end{array}$ & $\begin{array}{c} 0.08 \\ \mbf{0.66} \\ 0.26 \end{array}$ & $\begin{array}{c} 0.21 \\ \mbf{0.55} \\ 0.24 \end{array}$ & $\begin{array}{c} \mbf{0.45} \\ 0.36 \\ 0.19 \end{array}$ \\ \hline
F,M & $\begin{array}{c} \mbf{1.00} \\ 0.00 \\ 0.00 \end{array}$ & $\begin{array}{c} 0.00 \\ \mbf{1.00} \\ 0.00 \end{array}$ & $\begin{array}{c} 0.00 \\ \mbf{1.00} \\ 0.00 \end{array}$ & $\begin{array}{c} 0.03 \\ \mbf{0.82} \\ 0.15 \end{array}$ & $\begin{array}{c} 0.24 \\ 0.06 \\ \mbf{0.70} \end{array}$ & $\begin{array}{c} \mbf{0.74} \\ 0.26 \\ 0.00 \end{array}$ & $\begin{array}{c} \mbf{1.00} \\ 0.00 \\ 0.00 \end{array}$ & $\begin{array}{c} \mbf{1.00} \\ 0.00 \\ 0.00 \end{array}$ \\ \hline
F,N & $\begin{array}{c} \mbf{0.69} \\ 0.31 \\ 0.00 \end{array}$ & $\begin{array}{c} 0.23 \\ \mbf{0.52} \\ 0.25 \end{array}$ & $\begin{array}{c} 0.24 \\ \mbf{0.55} \\ 0.21 \end{array}$ & $\begin{array}{c} \mbf{0.94} \\ 0.03 \\ 0.03 \end{array}$ & $\begin{array}{c} \mbf{1.00} \\ 0.00 \\ 0.00 \end{array}$ & $\begin{array}{c} 0.26 \\ \mbf{0.59} \\ 0.15 \end{array}$ & $\begin{array}{c} 0.30 \\ \mbf{0.56} \\ 0.14 \end{array}$ & $\begin{array}{c} 0.37 \\ \mbf{0.49} \\ 0.14 \end{array}$ \\ \hline
F,O & $\begin{array}{c} \mbf{0.94} \\ 0.06 \\ 0.00 \end{array}$ & $\begin{array}{c} \mbf{0.48} \\ 0.40 \\ 0.12 \end{array}$ & $\begin{array}{c} \mbf{0.59} \\ 0.30 \\ 0.11 \end{array}$ & $\begin{array}{c} \mbf{0.99} \\ 0.01 \\ 0.00 \end{array}$ & $\begin{array}{c} \mbf{1.00} \\ 0.00 \\ 0.00 \end{array}$ & $\begin{array}{c} 0.12 \\ \mbf{0.66} \\ 0.22 \end{array}$ & $\begin{array}{c} 0.15 \\ \mbf{0.56} \\ 0.29 \end{array}$ & $\begin{array}{c} \mbf{0.40} \\ 0.31 \\ 0.29 \end{array}$ \\ \hline
F,G,H & $\begin{array}{c} 0.31 \\ \mbf{0.69} \\ 0.00 \end{array}$ & $\begin{array}{c} \mbf{0.89} \\ 0.02 \\ 0.09 \end{array}$ & $\begin{array}{c} \mbf{0.90} \\ 0.02 \\ 0.08 \end{array}$ & $\begin{array}{c} \mbf{0.53} \\ 0.17 \\ 0.30 \end{array}$ & $\begin{array}{c} \mbf{0.99} \\ 0.00 \\ 0.01 \end{array}$ & $\begin{array}{c} \mbf{0.73} \\ 0.18 \\ 0.09 \end{array}$ & $\begin{array}{c} 0.45 \\ \mbf{0.48} \\ 0.07 \end{array}$ & $\begin{array}{c} 0.23 \\ \mbf{0.69} \\ 0.08 \end{array}$ \\ \hline
F,J,M & $\begin{array}{c} \mbf{1.00} \\ 0.00 \\ 0.00 \end{array}$ & $\begin{array}{c} 0.00 \\ \mbf{1.00} \\ 0.00 \end{array}$ & $\begin{array}{c} 0.00 \\ \mbf{1.00} \\ 0.00 \end{array}$ & $\begin{array}{c} 0.07 \\ \mbf{0.87} \\ 0.06 \end{array}$ & $\begin{array}{c} 0.22 \\ 0.18 \\ \mbf{0.60} \end{array}$ & $\begin{array}{c} 0.00 \\ \mbf{0.98} \\ 0.02 \end{array}$ & $\begin{array}{c} 0.02 \\ \mbf{0.90} \\ 0.08 \end{array}$ & $\begin{array}{c} \mbf{0.68} \\ 0.21 \\ 0.11 \end{array}$ \\ \hline
\end{tabular}
\caption{Proportion of times each $K \in \{2,3,4\}$ (ordered from up $(K=2)$ to down $(K=4)$) was best (higher) in terms of the silhouette index for the models over $100$ generations of the SAR dataset with $100$ images per class.}
\label{tab:sar_unidimensional_silhouette_proportions}
\end{table}

\begin{table}[H]
\begin{tabular}{|c|c|c|c|c|c|c|c|c|c|c|}
\multicolumn{2}{@{}c@{}}{} & \multicolumn{1}{c}{Gaussian} & \multicolumn{1}{c}{Laplace} & \multicolumn{2}{@{}c@{}}{MG} & \multicolumn{3}{@{}c@{}}{Energy} \\
\multicolumn{1}{@{}c@{}}{Classes} & \multicolumn{1}{@{}c@{}}{2-W} & \multicolumn{1}{@{}c@{}}{\(\sigma^*\)} & \multicolumn{1}{@{}c@{}}{\(\sigma^*\)} & \multicolumn{1}{c}{$\alpha=2$} & \multicolumn{1}{c}{$\alpha=3$} & \multicolumn{1}{c}{$\alpha=0.25$} & \multicolumn{1}{c}{$\alpha=0.5$} & \multicolumn{1}{c}{$\alpha=0.75$} \\ \hline
F,G & $\begin{array}{c} 0.00 \\ 0.00 \\ \mbf{1.00} \end{array}$ & $\begin{array}{c} 0.00 \\ 0.00 \\ \mbf{1.00} \end{array}$ & $\begin{array}{c} 0.00 \\ 0.00 \\ \mbf{1.00} \end{array}$ & $\begin{array}{c} 0.00 \\ 0.00 \\ \mbf{1.00} \end{array}$ & $\begin{array}{c} 0.00 \\ 0.02 \\ \mbf{0.98} \end{array}$ & $\begin{array}{c} 0.00 \\ 0.00 \\ \mbf{1.00} \end{array}$ & $\begin{array}{c} 0.00 \\ 0.00 \\ \mbf{1.00} \end{array}$ & $\begin{array}{c} 0.00 \\ 0.00 \\ \mbf{1.00} \end{array}$ \\ \hline
F,H & $\begin{array}{c} 0.00 \\ 0.00 \\ \mbf{1.00} \end{array}$ & $\begin{array}{c} 0.00 \\ 0.00 \\ \mbf{1.00} \end{array}$ & $\begin{array}{c} 0.00 \\ 0.00 \\ \mbf{1.00} \end{array}$ & $\begin{array}{c} 0.00 \\ 0.02 \\ \mbf{0.98} \end{array}$ & $\begin{array}{c} 0.00 \\ 0.08 \\ \mbf{0.92} \end{array}$ & $\begin{array}{c} 0.00 \\ 0.00 \\ \mbf{1.00} \end{array}$ & $\begin{array}{c} 0.00 \\ 0.00 \\ \mbf{1.00} \end{array}$ & $\begin{array}{c} 0.00 \\ 0.00 \\ \mbf{1.00} \end{array}$ \\ \hline
F,I & $\begin{array}{c} 0.00 \\ 0.00 \\ \mbf{1.00} \end{array}$ & $\begin{array}{c} 0.00 \\ 0.00 \\ \mbf{1.00} \end{array}$ & $\begin{array}{c} 0.00 \\ 0.00 \\ \mbf{1.00} \end{array}$ & $\begin{array}{c} 0.00 \\ 0.02 \\ \mbf{0.98} \end{array}$ & $\begin{array}{c} 0.00 \\ 0.09 \\ \mbf{0.91} \end{array}$ & $\begin{array}{c} 0.00 \\ 0.00 \\ \mbf{1.00} \end{array}$ & $\begin{array}{c} 0.00 \\ 0.00 \\ \mbf{1.00} \end{array}$ & $\begin{array}{c} 0.00 \\ 0.00 \\ \mbf{1.00} \end{array}$ \\ \hline
F,J & $\begin{array}{c} 0.00 \\ 0.01 \\ \mbf{0.99} \end{array}$ & $\begin{array}{c} 0.00 \\ 0.00 \\ \mbf{1.00} \end{array}$ & $\begin{array}{c} 0.00 \\ 0.00 \\ \mbf{1.00} \end{array}$ & $\begin{array}{c} 0.00 \\ 0.05 \\ \mbf{0.95} \end{array}$ & $\begin{array}{c} 0.00 \\ 0.11 \\ \mbf{0.89} \end{array}$ & $\begin{array}{c} 0.00 \\ 0.00 \\ \mbf{1.00} \end{array}$ & $\begin{array}{c} 0.00 \\ 0.00 \\ \mbf{1.00} \end{array}$ & $\begin{array}{c} 0.00 \\ 0.00 \\ \mbf{1.00} \end{array}$ \\ \hline
F,K & $\begin{array}{c} 0.00 \\ 0.00 \\ \mbf{1.00} \end{array}$ & $\begin{array}{c} 0.00 \\ 0.00 \\ \mbf{1.00} \end{array}$ & $\begin{array}{c} 0.00 \\ 0.00 \\ \mbf{1.00} \end{array}$ & $\begin{array}{c} 0.00 \\ 0.02 \\ \mbf{0.98} \end{array}$ & $\begin{array}{c} 0.00 \\ 0.03 \\ \mbf{0.97} \end{array}$ & $\begin{array}{c} 0.00 \\ 0.00 \\ \mbf{1.00} \end{array}$ & $\begin{array}{c} 0.00 \\ 0.00 \\ \mbf{1.00} \end{array}$ & $\begin{array}{c} 0.00 \\ 0.00 \\ \mbf{1.00} \end{array}$ \\ \hline
F,L & $\begin{array}{c} 0.00 \\ 0.00 \\ \mbf{1.00} \end{array}$ & $\begin{array}{c} 0.00 \\ 0.00 \\ \mbf{1.00} \end{array}$ & $\begin{array}{c} 0.00 \\ 0.00 \\ \mbf{1.00} \end{array}$ & $\begin{array}{c} 0.00 \\ 0.00 \\ \mbf{1.00} \end{array}$ & $\begin{array}{c} 0.00 \\ 0.05 \\ \mbf{0.95} \end{array}$ & $\begin{array}{c} 0.00 \\ 0.00 \\ \mbf{1.00} \end{array}$ & $\begin{array}{c} 0.00 \\ 0.00 \\ \mbf{1.00} \end{array}$ & $\begin{array}{c} 0.00 \\ 0.00 \\ \mbf{1.00} \end{array}$ \\ \hline
F,M & $\begin{array}{c} 0.00 \\ 0.01 \\ \mbf{0.99} \end{array}$ & $\begin{array}{c} 0.00 \\ 0.00 \\ \mbf{1.00} \end{array}$ & $\begin{array}{c} 0.00 \\ 0.00 \\ \mbf{1.00} \end{array}$ & $\begin{array}{c} 0.00 \\ 0.05 \\ \mbf{0.95} \end{array}$ & $\begin{array}{c} 0.00 \\ 0.19 \\ \mbf{0.81} \end{array}$ & $\begin{array}{c} 0.00 \\ 0.00 \\ \mbf{1.00} \end{array}$ & $\begin{array}{c} 0.00 \\ 0.00 \\ \mbf{1.00} \end{array}$ & $\begin{array}{c} 0.00 \\ 0.00 \\ \mbf{1.00} \end{array}$ \\ \hline
F,N & $\begin{array}{c} 0.00 \\ 0.00 \\ \mbf{1.00} \end{array}$ & $\begin{array}{c} 0.00 \\ 0.00 \\ \mbf{1.00} \end{array}$ & $\begin{array}{c} 0.00 \\ 0.00 \\ \mbf{1.00} \end{array}$ & $\begin{array}{c} 0.00 \\ 0.00 \\ \mbf{1.00} \end{array}$ & $\begin{array}{c} 0.00 \\ 0.01 \\ \mbf{0.99} \end{array}$ & $\begin{array}{c} 0.00 \\ 0.00 \\ \mbf{1.00} \end{array}$ & $\begin{array}{c} 0.00 \\ 0.00 \\ \mbf{1.00} \end{array}$ & $\begin{array}{c} 0.00 \\ 0.00 \\ \mbf{1.00} \end{array}$ \\ \hline
F,O & $\begin{array}{c} 0.00 \\ 0.00 \\ \mbf{1.00} \end{array}$ & $\begin{array}{c} 0.00 \\ 0.00 \\ \mbf{1.00} \end{array}$ & $\begin{array}{c} 0.00 \\ 0.00 \\ \mbf{1.00} \end{array}$ & $\begin{array}{c} 0.00 \\ 0.02 \\ \mbf{0.98} \end{array}$ & $\begin{array}{c} 0.00 \\ 0.06 \\ \mbf{0.94} \end{array}$ & $\begin{array}{c} 0.00 \\ 0.00 \\ \mbf{1.00} \end{array}$ & $\begin{array}{c} 0.00 \\ 0.00 \\ \mbf{1.00} \end{array}$ & $\begin{array}{c} 0.00 \\ 0.00 \\ \mbf{1.00} \end{array}$ \\ \hline
F,G,H & $\begin{array}{c} 0.00 \\ 0.00 \\ \mbf{1.00} \end{array}$ & $\begin{array}{c} 0.00 \\ 0.00 \\ \mbf{1.00} \end{array}$ & $\begin{array}{c} 0.00 \\ 0.00 \\ \mbf{1.00} \end{array}$ & $\begin{array}{c} 0.00 \\ 0.00 \\ \mbf{1.00} \end{array}$ & $\begin{array}{c} 0.00 \\ 0.02 \\ \mbf{0.98} \end{array}$ & $\begin{array}{c} 0.00 \\ 0.00 \\ \mbf{1.00} \end{array}$ & $\begin{array}{c} 0.00 \\ 0.00 \\ \mbf{1.00} \end{array}$ & $\begin{array}{c} 0.00 \\ 0.00 \\ \mbf{1.00} \end{array}$ \\ \hline
F,J,M & $\begin{array}{c} 0.00 \\ 0.00 \\ \mbf{1.00} \end{array}$ & $\begin{array}{c} 0.00 \\ 0.00 \\ \mbf{1.00} \end{array}$ & $\begin{array}{c} 0.00 \\ 0.00 \\ \mbf{1.00} \end{array}$ & $\begin{array}{c} 0.00 \\ 0.05 \\ \mbf{0.95} \end{array}$ & $\begin{array}{c} 0.00 \\ 0.22 \\ \mbf{0.78} \end{array}$ & $\begin{array}{c} 0.00 \\ 0.00 \\ \mbf{1.00} \end{array}$ & $\begin{array}{c} 0.00 \\ 0.00 \\ \mbf{1.00} \end{array}$ & $\begin{array}{c} 0.00 \\ 0.00 \\ \mbf{1.00} \end{array}$ \\ \hline
\end{tabular}
\caption{Proportion of times each $K \in \{2,3,4\}$ (ordered from up $(K=2)$ to down $(K=4)$) was best (lowest) in terms of the Davies-Bouldin* index for the models over $100$ generations of the SAR dataset with $100$ images per class.}
\label{tab:sar_unidimensional_davies_bouldin_star_proportions}
\end{table}


\section{Conclusions} \label{Section.Conclusions}

In this work we have proposed a kernel \(K\)-means clustering procedure for distributional data defined on \(\bbR^p\). Given a kernel \(k\) and its associated RKHS \(\scrH\), the sampled distributions \(F_i\), \(i=1,\ldots,n\), are (implicitly) mapped into their corresponding kernel mean embedding \(\mu_{F_i}\in\scrH\), \(i=1,\ldots,n\). Then this latter sample is clustered via \(K\)-means with respect to the maximum mean discrepancy metric in \(\scrH\). The method is simple to understand and implement. The behaviour of this clustering procedure has been assessed on simulated and real one- and two-dimensional distributional data and with different kernels. The results are satisfactory, although they strongly depend on the choice of the kernel and, to a lesser extent, on its tuning parameter. The energy kernel appears as a safe choice. The proposed kernel \(K\)-means methodology provides a non-hierarchical clustering approach in the setting of two-dimensional distributions where previous proposals in the literature worked only under considerably more limited assumptions. Future research can focus on the choice of the tuning parameter of the kernel and the scalability for samples with large size \(n\) or high dimensionalily (large \(p\)).


\section*{Acknowledgements}

A.B. and J.R.B. are supported by the Spanish MCyT grant PID2023-148081NB-I00. J.R.B. acknowledges financial support from Grant CEX2023-001347-S funded by MICIU/AEI/10.13039/501100011033.



\begin{thebibliography}{99}
\bibitem[\protect\citeauthoryear{Applegate {\em et al.}}{2011}]{Applegate_et_al_11} Applegate, D.,
Dasu, T., Krishnan, S. and Urbanek, S. (2011). Unsupervised clustering of multidimensional distributions using earth mover distance. In {\em KDD '11: Proceedings of the 17th ACM SIGKDD international conference on Knowledge discovery and data mining}, 636--644. 
\bibitem[\protect\citeauthoryear{Arbelaitz {\em et al.}}{2013}]{Arbelaitz_et_al_13} Arbelaitz, O., Gurrutxaga, I., Muguerza, J., P\'{e}rez, J.M. and Perona, I. (2013). An extensive comparative study of cluster validity indices. {\em Pattern Recognition}, 46, 243--256. 
\bibitem[\protect\citeauthoryear{Arias-Castro and Qiao}{2025}]{AriasCastroQiao25} Arias--Castro, E. and Qiao, W. (2025). Embedding distributional data. {\em The Annals of Statistics}, to appear. 
\bibitem[\protect\citeauthoryear{del Barrio {\em et al.}}{2019}]{Barrio_et_al_19} del Barrio, E., Cuesta-Albertos, J.A., Matr\'{a}n, C. and Mayo--Iscar, A. (2019). Robust clustering tools based on optimal transportation. {\em Statistics and Computing}, 29, 139–-160. (2019). 
\bibitem[\protect\citeauthoryear{Bigot}{2020}]{Bigot20} Bigot, J. (2020). Statistical data analysis in the Wasserstein space. {\em ESAIM: Proceedings and Surveys}, 68, 1--19.
\bibitem[\protect\citeauthoryear{Billard}{2006}]{Billard06} Billard, L. (2006). Symbolic data analysis: what is it?. In: Rizzi, A., Vichi, M. (eds) {\em Compstat 2006 - Proceedings in Computational Statistics}, pp. 261--269. Physica-Verlag. 
\bibitem[\protect\citeauthoryear{Brito and Dias}{2022}]{BritoDias22} Brito, P. and Dias,S. (2022). {\em Analysis of Distributional Data}. CRC Press.
\bibitem[\protect\citeauthoryear{Caliński and Harabasz}{1974}]{CalinskiHarabasz74} Caliński, T. and  Harabasz, J. (1974). A dendrite method for cluster analysis. {\em Communications in Statistics: Theory and Methods}, 3(1), 1–-27.
\bibitem[\protect\citeauthoryear{Chen and Hung}{2014}]{ChenHung14} Chen, J.-H. and Hung, W.-L. (2014). An automatic clustering algorithm for probability density functions. {\em Journal of Statistical Computation and Simulation}, 85(15), 3047--3063. 
\bibitem[\protect\citeauthoryear{Dhillon {\em et al.}}{2004}]{Dhillon_et_al_04} Dhillon, I., Guan, Y. and Kulis, B. (2004). Kernel k-means, Spectral Clustering and Normalized Cuts. In {\em KDD '04: Proceedings of the tenth ACM SIGKDD international conference on Knowledge discovery and data mining}, 551--556. 
\bibitem[\protect\citeauthoryear{Gachon {\em et al.}}{2025}]{Gachon_et_al_25} Gachon, E., Bigot, J. and  Cazelles, E. (2025). Scalable and consistent embedding of probability measures into Hilbert spaces via measure quantization. arXiv:2502.04907v1
\bibitem[\protect\citeauthoryear{Goh and Vidal}{2008}]{GohVidal08} Goh, A. and Vidal, R. (2008). Unsupervised Riemannian clustering of probability density functions. In: Daelemans, W., Goethals, B., Morik, K. (eds) {\em Machine Learning and Knowledge Discovery in Databases}. ECML PKDD 2008. Lecture Notes in Computer Science, vol. 5211. Springer. 
\bibitem[\protect\citeauthoryear{Gretton {\em et al.}}{2012a}]{Gretton_et_al_12a} Gretton, A., Borgwardt, K., Rasch,  M., Sch\"{o}lkopf, B. and Smola,  A. (2012a). A kernel two-sample test. {\em Journal of Machine Learning Research}, 13, 723--773. 
\bibitem[\protect\citeauthoryear{Gretton {\em et al.}}{2012b}]{Gretton_et_al_12b}  Gretton, A., Sriperumbudur,  B., Sejdinovic, D., Strathmann, H., Balakrishnan, S., Pontil, M. and Fukumizu, K. (2012b). Optimal kernel choice for large-scale two-sample tests. In {\em Advances in Neural Information Processing Systems}, pp. 1214–-1222.
\bibitem[\protect\citeauthoryear{Henderson {\em et al.}}{2015}]{EP_MEANS15} Henderson, K., Gallagher, B. and Eliassi-Rad, T. (2015). EP-MEANS: an efficient nonparametric clustering of empirical probability distributions. {\em SAC '15: Proceedings of the 30th Annual ACM Symposium on Applied Computing}, pp. 893--900. 
\bibitem[\protect\citeauthoryear{Hubert and Arabie}{1985}]{HubertArabie85} Hubert, L. and Arabie, P. (1985). Comparing partitions. {\em Journal of Classification}, 2, 193--218.
\bibitem[\protect\citeauthoryear{Irpino and Verde}{2006}]{IrpinoVerde06} Irpino, A. and Verde, R. (2006). A new Wasserstein based distance for the hierarchical clustering of histogram symbolic data. In: Batagelj, V., Bock, HH., Ferligoj, A., \v{Z}iberna, A. (eds) Data Science and Classification. Studies in {\em Classification, Data Analysis, and Knowledge Organization}. Springer. 
\bibitem[\protect\citeauthoryear{Irpino {\em et al.}}{2014}]{Irpino_et_al_14} Irpino, A., Verde, R. and de Carvalho, F.A.T. (2014). Dynamic clustering of histogram data based on adaptive squared Wasserstein distances. {\em Expert Systems with Applications}, 41, 3351--3366.
\bibitem[\protect\citeauthoryear{Irpino {\em et al.}}{2017}]{Irpino_et_al_17} Irpino, A., Verde, R.  and de Carvalho, F.A.T. (2017). Fuzzy clustering of distributional data with automatic weighting of variable components. {\em Information Sciences}, 406, 248--268.
\bibitem[\protect\citeauthoryear{Jacques and Preda}{2014}]{JacquesPreda14} Jacques, J. and Preda, C. (2014). Functional data clustering: a survey. {\em Advances in Data Analysis and Classification}, 8, 231–-255. 
\bibitem[\protect\citeauthoryear{Kim and Ramakrishna}{2005}]{KimRamakrishna05} Kim, M. and Ramakrishna, R.S. (2005). New indices for cluster validity assessment. {\em Pattern Recognition Letters}, 26, 2353-–2363.
\bibitem[\protect\citeauthoryear{Li {\em et al.}}{2016}]{Li_et_al_16} Li, H.--C., Krylov, V.A., Fan, P.--Z., Zerubia, J. and Emery, W.J. (2016). Unsupervised learning of generalized gamma mixture model with application in statistical modeling of high-resolution SAR images. {\em IEEE Transactions on Geoscience and Remote Sensing}, 54, 4, 2153--2170. 
\bibitem[\protect\citeauthoryear{Meester and Baslamisli}{2022}]{meester2022sar} Meester, M. J. and Baslamisli, A. S. (2022). SAR image edge detection: review and benchmark experiments. {\em International Journal of Remote Sensing}, 43, 14, 5372--5438. 
\bibitem[\protect\citeauthoryear{Modeste and Dombry}{2024}]{ModesteDombry24} Modeste, T. and Dombry, C. (2024). Characterization of translation invariant MMD on \(\bbR^d\) and connections with Wasserstein distances. {\em Journal of Machine Learning Research}, 25(237), 1--39. 
\bibitem[\protect\citeauthoryear{Montanari and Cal\`{o}}{2013}]{MontanariCalo13} Montanari, A. and Cal\`{o}, D. (2013). Model-based clustering of probability density functions. {\em Advances in Data Analysis and Classification}, 7(3), 301–-319. 
\bibitem[\protect\citeauthoryear{Mora--L\'{o}pez and Mora}{2015}]{MoraLopezMora15} Mora--L\'{o}pez, L. and Mora, J. (2015). An adaptive algorithm for clustering cumulative probability distribution functions using the Kolmogorov--Smirnov two--sample test. {\em Expert Systems with Applications}, 42, 4016--4021.
\bibitem[\protect\citeauthoryear{Moreira {\em et al.}}{2013}]{Moreira_etal_13} Moreira, A., Prats--Iraola, P., Younis, M., Krieger, G., Hajnsek, I. and Papathanassiou, K.P. (2013). A tutorial on synthetic aperture radar. {\em IEEE Geoscience and Remote Sensing Magazine}, 1(1), 6–-43.
\bibitem[\protect\citeauthoryear{Muandet {\em et al.}}{2017}]{BOOK_KernelMeanEmbedDistrib} Muandet, K., Fukumizu, K., Sriperumbudur, B. and Sch\"{o}lkopf, B. (2017). Kernel Mean Embedding of Distributions: A Review and Beyond. {\em Foundations and Trends in Machine Learning}, 10, 1--141. 
\bibitem[\protect\citeauthoryear{Okano and Imaizumi}{2024}]{OkanoImaizumi24} Okano, R. and Imazumi, M. (2024). Wasserstein k-centers clustering for distributional data. arXiv:2407.08228v3.
\bibitem[\protect\citeauthoryear{Panaretos and Zemel}{2019}]{PanaretosZemel19} Panaretos, V.M. and Zemel, Y. (2019). Statistical aspects of Wasserstein distances. {\em Annual Review of Statistics and Its Application}, 6, 405--431.
\bibitem[\protect\citeauthoryear{Petersen and M\"{u}ller}{2016}]{PetersenMuller16} Petersen, A. and M\"{u}ller, H.--G. (2016). Functional data analysis for density functions by transformation to a Hilbert space. {\em The Annals of Statistics}, 44, 183–-218.
\bibitem[\protect\citeauthoryear{Petersen {\em et al.}}{2022}]{DensitiesAsDataObjects} Petersen, A., Zhang, C. and Kokoszka, P. (2022). Modeling probability density functions as data objects. {\em Econometrics and Statistics}, 21, 159--178. 
\bibitem[\protect\citeauthoryear{Rousseeuw}{1987}]{Rousseeuw87} Rousseeuw, P.J. (1987). Silhouettes: a graphical aid to the interpretation and validation of cluster analysis. {\em Journal of Computational and Applied Mathematics}, 20, 53--65.
\bibitem[\protect\citeauthoryear{Sejdinovic {\em et al.}}{2013}]{Sejdinovic_etal_13} Sejdinovic, D., Sriperumbudur, B., Gretton, A. and Fukumizu, K. (2013). Equivalence of distance-based and RKHS-based statistics in hypothesis testing. {\em The Annals of Statistics}, 41(5), 2263--2291.
\bibitem[\protect\citeauthoryear{Smola {\em et al.}}{2007}]{HilbSpEmbedDistrib} Smola, A., Gretton, A., Song, L. and  Sch\"{o}lkopf, B. (2007). A Hilbert space embedding for distributions. In {\em Proceedings of the 18th International Conference on Algorithmic Learning Theory}, pp. 13--31. Springer-Verlag.
\bibitem[\protect\citeauthoryear{Sriperumbudur {\em et al.}}{2009}]{Sriperumbudur_etal_09} Sriperumbudur, B., Fukumizu, K., Gretton, A. Lanckriet, G. and Sch\"{o}lkopf, B. (2009). Kernel choice and classifiability for RKHS embeddings of probability distributions. In {\em Advances in Neural Information Processing Systems} 22, pp. 1750–-1758.
\bibitem[\protect\citeauthoryear{Szekely and Rizzo}{2023}]{SzekelyRizzo23} Sz\'{e}kely, G. and Rizzo, M. (2023). {\em The Energy of Data and Distance Correlation}. CRC Press.
\bibitem[\protect\citeauthoryear{Vega--Pons and Ruiz--Shulcloper}{2011}]{VegaPRuizS11} Vega--Pons, S. and Ruiz--Shulcloper, J. (2011). A survey of clustering ensemble algorithms. {\em International Journal of Pattern Recognition and Artificial Intelligence}, 25(3), 337--372.
\bibitem[\protect\citeauthoryear{Verdinelli and Wasserman}{2019}]{VerdinelliWasserman19} Verdinelli, I. and Wasserman, L. (2019). Hybrid Wasserstein distance and fast distribution clustering. {\em Electronic Journal of Statistics}, 13(2), 5088--5119.
\bibitem[\protect\citeauthoryear{Vo Van and Pham--Gia}{2010}]{VovanPhamgia10} Vo Van, T. and Pham--Gia,T. (2010). Clustering probability distributions. {\em Journal of Applied Statistics}, 37(11), 1891-–1910. 
\bibitem[\protect\citeauthoryear{Vrac {\em et al.}}{2012}]{Vrac_et_al_12} Vrac, M., Billard, L., Diday,  E. and Ch\'{e}din, A. (2012). Copula analysis of mixture models. {\em Computational Statistics}, 27, 427--457. 
\bibitem[\protect\citeauthoryear{Wang {\em et al.}}{2018}]{Wang_et_al_18} Wang, C., Mouche, A., Tandeo, P., Stopa, J., Long\'{e}p\'{e}, N., Erhard, G., Foster, R., Vandemark, D. and Chapron, B. (2018). Labeled SAR imagery dataset of ten geophysical phenomena from Sentinel-1 wave mode (TenGeoP-SARwv). {\em SEANOE}, \url{https://www.seanoe.org/data/00456/56796/}. 
\bibitem[\protect\citeauthoryear{Wang {\em et al.}}{2019}]{Wang_et_al_19} Wang, C., Mouche, A., Tandeo, P., Stopa, J., Long\'{e}p\'{e}, N., Erhard, G., Foster, R., Vandemark, D. and Chapron, B. (2019). A labelled ocean SAR imagery dataset of ten geophysical phenomena from Sentinel‐1 wave mode. {\em Geoscience Data Journal}, 6, 2, 105--115. 
\bibitem[\protect\citeauthoryear{Zambom {\em et al.}}{2022}]{Zambom_et_al_22} Zambom, A.Z., Collazos, J.A. and Dias, R. (2022). Selection of the number of clusters in functional data analysis. {\em Journal of Statistical Computation and Simulation}, 92(14), 2980-–2998. 
\bibitem[\protect\citeauthoryear{Zhu {\em et al.}}{2021}]{Zhu21BankCard} Zhu, Y., Deng, Q., Huang, D.,  Jing, B. and Zhang, B. (2021). Clustering based on Kolmogorov–Smirnov statistic with application to bank card transaction data. {\em Journal of the Royal Statistical Society Series C: Applied Statistics}, 70 (3), 558–-578. 
\end{thebibliography}
\end{document}